\def\paperlanguage{} 
\pdfoutput=1 

\documentclass[letterpaper, 10 pt, conference]{ieeeconf}  

\usepackage{bm}
\usepackage{cite}
\usepackage{flushend}
\usepackage{here}
\usepackage{multirow}
\include{preamble}

\IEEEoverridecommandlockouts                              

\title{
  {
    \switchlanguage%
    {%
    SAQIEL: Ultra-Light and Safe Manipulator\\with Passive 3D Wire Alignment Mechanism 
    }%
    {%
    SAQIEL: Ultra-Light and Safe Manipulator\\with Passive 3D Wire Alignment Mechanism 
    }%
  }
}

\author{Temma Suzuki$^{1}$, Masahiro Bando$^{1}$, Kento Kawaharazuka$^{1}$, Kei Okada$^{1}$, and Masayuki Inaba$^{1}$
  \thanks{
    $^{1}$ The authors are with the Department of Mechano-Informatics, Graduate School of Information Science and Technology, The University of Tokyo, 7-3-1 Hongo, Bunkyo-ku, Tokyo, 113-8656, Japan.
    {\text\small [suzuki, bando, kawaharazuka, okada, inaba]@jsk.t.u-tokyo.ac.jp}
  }
}
\begin{document}

\maketitle

\begin{abstract}
  \switchlanguage%
  {%
    Improving the safety of collaborative manipulators necessitates the reduction of inertia in the moving part. 
    Within this paper, we introduce a novel approach in the form of a passive 3D wire aligner, serving as a lightweight and low-friction power transmission mechanism, thus achieving the desired low inertia in the manipulator's operation. 
    Through the utilization of this innovation, the consolidation of hefty actuators onto the root link becomes feasible, consequently enabling a supple drive characterized by minimal friction.
    To demonstrate the efficacy of this device, we fabricate an ultralight 7 degrees of freedom (DoF) manipulator named SAQIEL, boasting a mere \SI{1.5}{\kilogram} weight for its moving components. Notably, to mitigate friction within SAQIEL's actuation system, we employ a distinctive mechanism that directly winds wires using motors, obviating the need for traditional gear or belt-based speed reduction mechanisms. 
    Through a series of empirical trials, we substantiate that SAQIEL adeptly strikes balance between lightweight design, substantial payload capacity, elevated velocity, precision, and adaptability.
  }%
  {%
    協働マニュピレータの安全性向上には動作部の低慣性化が重要である.
    本論文ではマニュピレータの低慣性化のための軽量・低摩擦な動力伝達機構として, 受動3次元ワイヤ整列装置を提案する.
    本装置を用いることで重いアクチュエータや減速機をルートリンクに集約しながら, 低摩擦で柔軟な駆動を実現可能である.
    この装置の有効性を示すため, 動作部重量\SI{1.5}{\kilogram}の超軽量7自由度マニュピレータSAQIELを作成する.
    さらにSAQIELでは駆動系の低摩擦化のため, ギアやベルトによる減速機構を使用せずモータでワイヤを直接巻き取る機構を採用する.
    そしてSAQIELが軽量性とペイロード・高速性・正確性・柔軟性を両立できることを実機実験を通して示す.
  }%
\end{abstract}


\section{Introduction}\label{sec:introduction}
\switchlanguage%
{%
In recent years, with the expansion of robots into society, there is a growing demand for manipulators that can operate in the same environments as humans.
While conventional industrial robots avoid unintended collisions with their environment by operating within enclosures, manipulators designed to operate in living environments have a higher likelihood of unintended collisions with humans and surroundings. 
For such manipulators, innovative solutions are required to prevent damage to both the robot and the environment during collisions.

A quantifiable measure to assess robot safety is the maximum contact force when a human and a robot come into contact. 
When the human body's contact points are not constrained by walls or floors, the maximum external contact force $F^{\mathrm{max}}_{\mathrm{ext}}$ is expressed by the following equation \cite{Haddadin2015}:
\begin{align}
  \label{eq:fext_max}
  F^{\mathrm{\max}}_{\mathrm{ext}}=\sqrt{\frac{m_\mathrm{u} M_\mathrm{H}}{m_\mathrm{u}+M_\mathrm{H}}} \sqrt{K_\mathrm{H}} \dot{x}_{\mathrm{re}}^0
\end{align}
here, $m_\mathrm{u}$ represents the effective mass of the robot's contact point, $M_\mathrm{H}$ is the effective mass of the human body's contact point, $K_\mathrm{H}$ stands for the effective stiffness of the contact point, and $\dot{x}_{\mathrm{re}}^0$ denotes the relative collision velocity between the robot and the human. 
From this equation, it is apparent that reducing the effective mass $m_\mathrm{u}$ of the robot can decrease the maximum contact force $F^{\mathrm{max}}_{\mathrm{ext}}$ without lowering the collision speed $\dot{x}_{\mathrm{re}}^0$. 
Thus, efforts have been made to achieve both high-speed operation and safety by minimizing the inertia of the manipulator's moving part.

\begin{figure}[t]
  \centering
  \includegraphics[width=0.9\columnwidth]{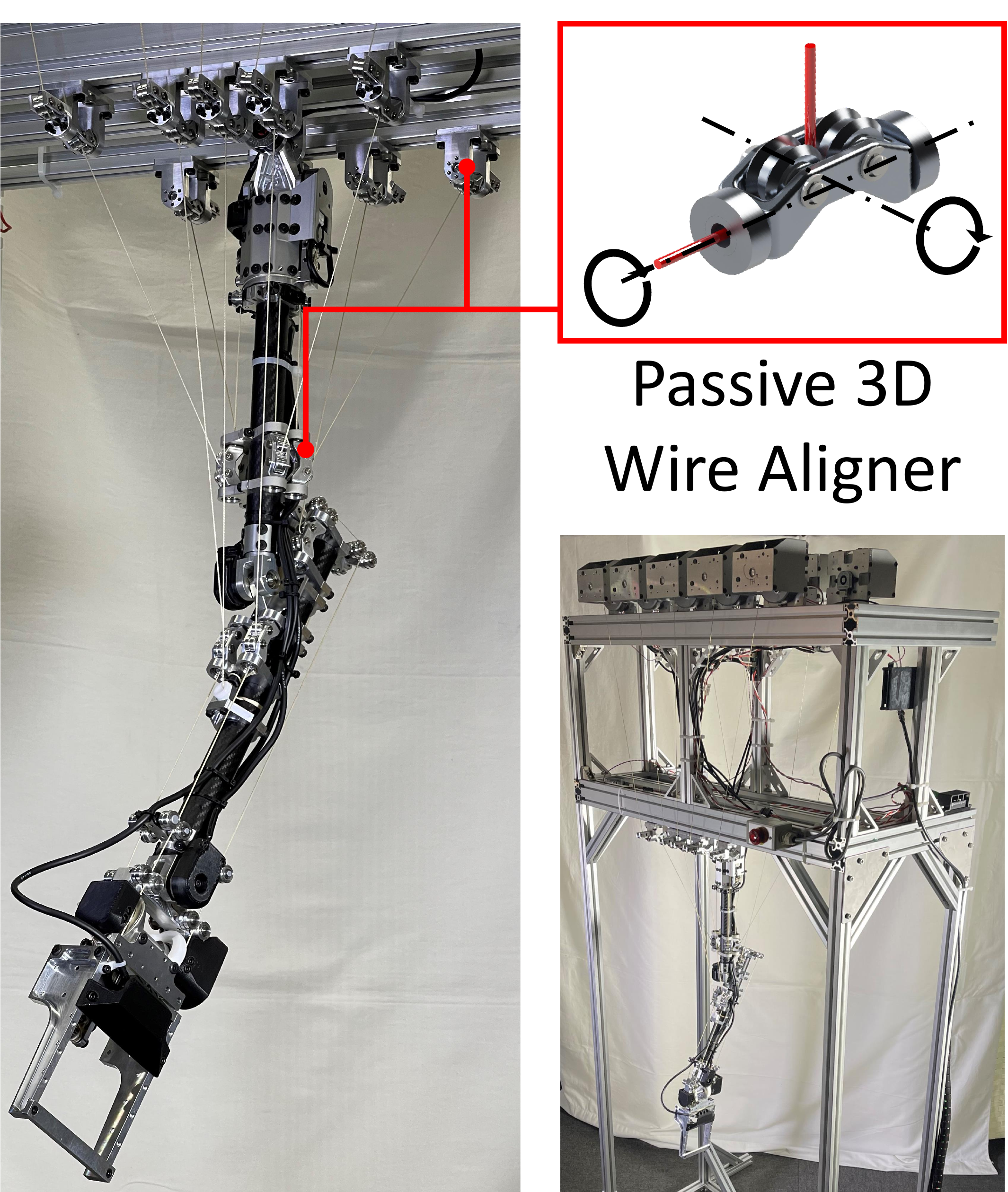}
  \vspace{-1ex}
  \caption{Overview of passive 3D wire aligner and 7-DoF manipulator SAQIEL with this mechanism.}
  \label{figure:wirealigneroverview}
  \vspace{-1.5ex}
\end{figure}

Among the components comprising a manipulator, the gearbox and actuators contribute significantly to the overall weight. 
Kim and colleagues have approached the challenge by utilizing wires for power transmission, consolidating motors that drive the manipulator's wrist and elbow onto the upper arm, thus reducing the inertia of the moving part \cite{kim2017anthropomorphic, song2018lims2, lims32022}. 
However, a drawback of this method is that actuators are still mounted on the moving part.

A method to aggregate motors on the root link is the concept of coupled tendon-driven actuation \cite{hirose1991tendon, yokoi1991design}. 
This technique employs a redundant number of wires relative to the number of joints to transmit power. 
As a result, actuators can be completely removed from the moving part. 
The coupled tendon-driven robot arm developed by Yokoi et al. had a total length of \SI{843}{\milli\metre} and a moving part weight of \SI{4}{\kilogram} \cite{yokoi1991design}. 
Considering that a commercially available collaborative robot arm of similar size (total length \SI{850}{\milli\metre}) has a moving part weight of \SI{17.7}{\kilogram} \cite{ur5e2023}, it becomes evident that coupled tendon-driven actuation can significantly reduce the moving part weight.

However, in coupled tendon-driven actuation, since the wires pass through all joints up to the terminal link, a large-diameter pulley wound with wires is necessary for each joint (about half the number of wires times the number of joints). 
Consequently, the complexity of the joint mechanism increases, and there is a potential drawback of increased overall weight of the moving part.

As a simple and lightweight method to transmit power from actuators on the root link to each joint, one approach is to use wires and tubes.
In this method, wires pass through tubes to transmit power from the root link to the desired link \cite{mustafa2008self, chen2013integrated}. 
Using this approach, Mustafa et al. achieved a lightweight 7-DoF robot arm with a moving part weight of \SI{1}{\kilogram} \cite{mustafa2008self}.
This approach yields a simple and lightweight mechanism; however, it introduces challenges such as friction between the wires and tubes, making the modeling of the wire-driven system complex.

Therefore, in this study, we aim to create a lightweight, low-friction mechanism that enables power transmission across multiple joints and works toward achieving a safe multi-degree-of-freedom manipulator. 
To achieve this goal, we propose a passive 3D wire aligner, as illustrated in \figref{figure:wirealigneroverview}, which allows transmission of wires across multiple joints.

It should be noted that the primary focus of this study is on the hardware design of a lightweight, flexible and safe manipulator, rather than the improvement of end-effector positioning accuracy or the development of new control methods for wire-driven robots.

\begin{figure}[t]
  \centering
  \includegraphics[width=1.0\columnwidth]{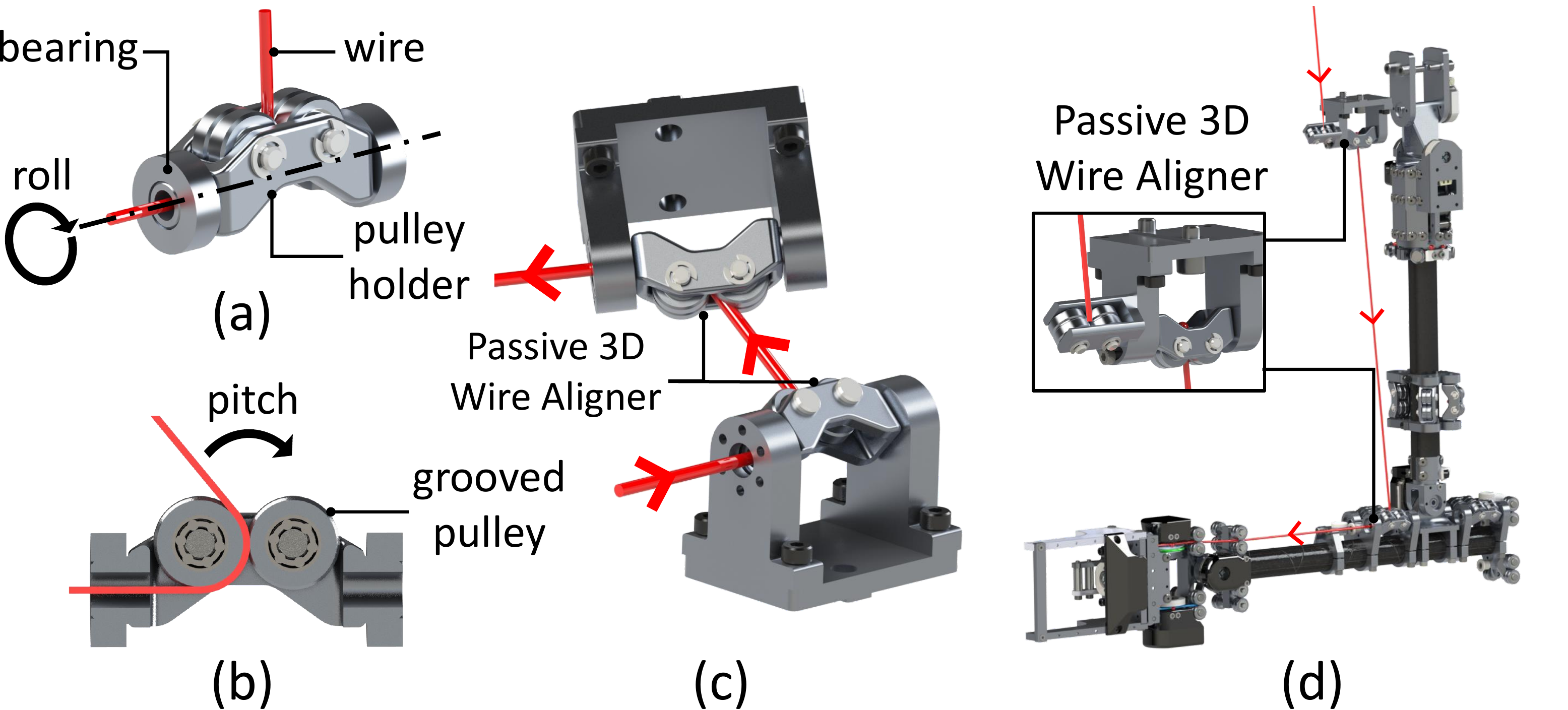}
  \vspace{-2ex}
  \caption{Detail of passive 3D wire aligner. (a) overview of the wire aligner and roll axis rotation of the wire, (b) cross section of the wire aligner and pitch axis rotation of the wire, (c) pair of wire aligner, (d) wire aligner transmitting a wire inside a 7-DoF manipulator.}
  \label{figure:wirealignerstructure}
  \vspace{-1.5ex}
\end{figure}
}%
{

近年ロボットの社会進出に伴い, 人間と同じ環境で動作するマニュピレータの需要が高まっている.
一般的な産業用ロボットは柵の中で作業することで環境との意図しない衝突を回避している.
一方で生活環境で動作するマニュピレータは, 人間や環境に対して意図しない衝突をする可能性が高い.
このようなマニュピレータには外部との衝突時にロボットと環境双方が損傷しないための工夫が求められる.

ロボットの安全性を定量的に評価する指標として, 人間とロボットが接触した際の最大接触力が有効である. 
人体の接触部が壁や床によって拘束されていないとき, 最大接触力$F^{\mathrm{\max}}_{\mathrm{ext}}$は以下の式で表される\cite{Haddadin2015}.

\begin{align}
  \label{eq:fext_max}
  F^{\mathrm{\max}}_{\mathrm{ext}}=\sqrt{\frac{m_\mathrm{u} M_\mathrm{H}}{m_\mathrm{u}+M_\mathrm{H}}} \sqrt{K_\mathrm{H}} \dot{x}_{\mathrm{re}}^0
\end{align}

ただし, $m_\mathrm{u}$はロボットの接触部の有効質量, $M_\mathrm{H}$は人体の接触部の有効質量, $K_\mathrm{H}$は接触部の実効剛性, $\dot{x}_{\mathrm{re}}^0$はロボットと人間の相対衝突速度である.
上式からロボットの有効質量$m_\mathrm{u}$を小さくすることで動作速度$\dot{x}_{\mathrm{re}}^0$を低下させずに最大接触力$F^{\mathrm{\max}}_{\mathrm{ext}}$を低減可能であることが分かる.
このためマニュピレータ動作部の慣性を小さくすることで高速動作と安全性を両立する試みが行われている.

\begin{figure}[t]
  \centering
  \includegraphics[width=1.0\columnwidth]{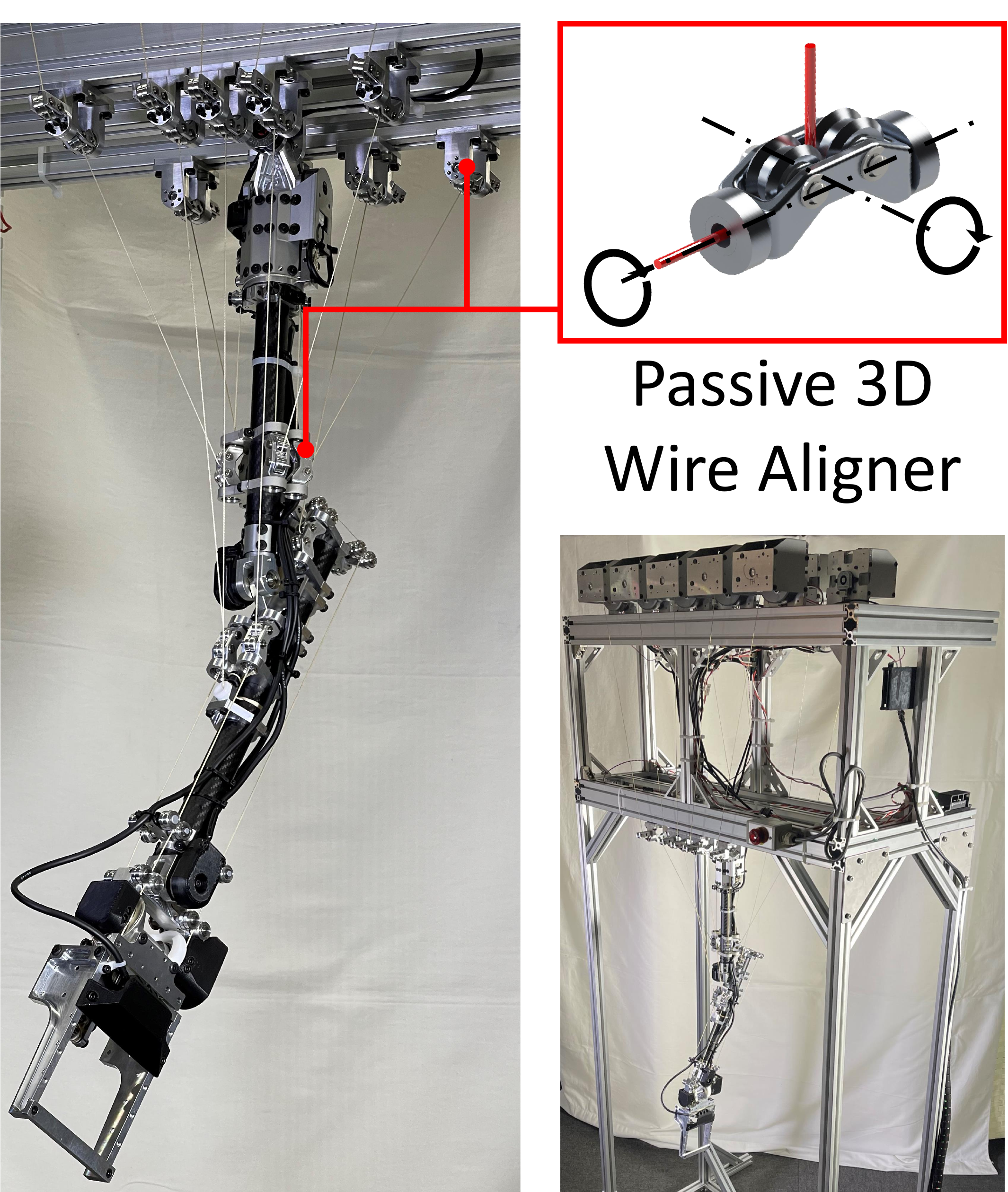}
  \vspace{-2ex}
  \caption{Overview of passive 3D wire aligner and 7-DoF manipulator SAQIEL with this mechanism.}
  \label{figure:wirealigneroverview}
  \vspace{-1.5ex}
\end{figure}

\begin{figure}[H]
  \centering
  \includegraphics[width=1.0\columnwidth]{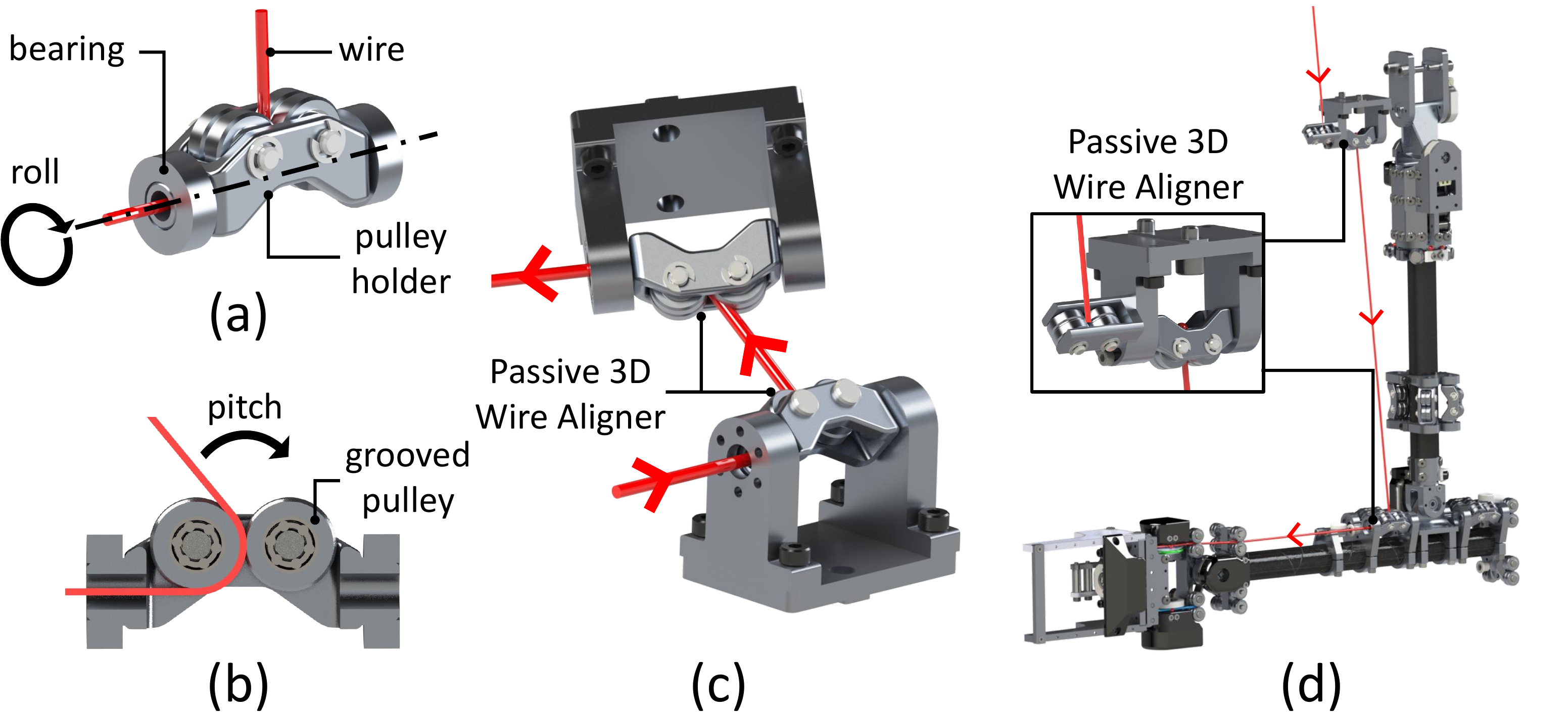}
  \vspace{-2ex}
  \caption{Detail of passive 3D wire aligner. (a) overview of the wire aligner and roll axis rotation of the wire, (b) cross section of the wire aligner and pitch axis rotation of the wire, (c) pair of wire aligner, (d) wire aligner transmitting a wire inside a 7-DoF manipulator.}
  \label{figure:wirealignerstructure}
  \vspace{-1.5ex}
\end{figure}


マニュピレータの構成要素の中で大きな重量比を占めるのは減速機とアクチュエータである.
Kimらは, ワイヤを動力伝達に用いることでマニュピレータの手首や肘を駆動するモータを上腕に集約し, 動作部慣性を小さくするアプローチを取っている \cite{kim2017anthropomorphic, song2018lims2, lims32022}.
一方でこの手法では動作部からモータを排除しきれていないという問題点がある.

モータをルートリンクに集約する方法としてはワイヤ干渉駆動が挙げられる \cite{hirose1991tendon, yokoi1991design}.
ワイヤ干渉駆動では関節数に対して冗長な本数のワイヤを用いて動力を伝達する.
そのため動作部からアクチュエータを完全に排除できるという利点がある.
yokoiらの開発したワイヤ干渉駆動ロボットアームは全長\SI{843}{\milli\metre}で動作部重量\SI{4}{\kilogram}であった\cite{yokoi1991design}.
同等なサイズ(全長\SI{850}{\milli\metre})の市販のロボットアームが動作部重量\SI{17.7}{\kilogram}であることを考えると, ワイヤ干渉駆動によって動作部重量を大幅に軽減できることが分かる\cite{ur5e2023}.

一方でワイヤ干渉駆動ではワイヤが終端リンクまでの全ての関節を通過するため, ワイヤを巻きつける大径プーリーが関節数$\times$ワイヤ本数/2程度必要となる.
これにより関節機構が複雑化しやすく, 動作部の全体重量が増加しうるという欠点がある.

ルートリンク上のアクチュエータの動力を各関節に伝達する簡易・軽量な手法としては, ワイヤとチューブを用いるものがある.
この方式ではワイヤをチューブに通し, ルートリンクから所望のリンクまで動力を伝達する\cite{mustafa2008self, chen2013integrated}.
Mustafaらはこの手法を用いて, 動作部重量\SI{1}{\kilogram}の軽量な7自由度ロボットアームを実現した\cite{mustafa2008self}.
このアプローチは機構が簡便で軽量な一方で, ワイヤとチューブが摺動しワイヤ駆動系にモデル化困難な摩擦が発生しうるという問題点がある.

そこで本研究では複数関節をまたいだワイヤの伝達を可能とする軽量・低摩擦な機構を作成し, 安全な多自由度マニュピレータの実現を目指す.
そしてこの目的を満たす手法として, \figref{figure:wirealigneroverview}に示すような受動3次元ワイヤ整列装置を提案する.

なお本研究は手先位置決め精度の向上やワイヤ駆動ロボット用の新しい制御手法の開発よりも, 軽量柔軟で安全なマニュピレータのハードウェア設計に焦点を当てていることに注意されたい.

}%

\section{Passive 3D Wire Aligner} \label{sec:aligner}
\subsection{Feature of Passive 3D Wire Aligner} \label{subsec:aligner-feature}
\switchlanguage%
{%
The detailed structure of the passive 3D wire aligner is illustrated in \figref{figure:wirealignerstructure}. 
When a wire passes through this device, it aligns with a predetermined straight line regardless of the angle at which the wire enters the device.
By installing a wire aligner on each of the two links, a wire path connecting the two links in a straight line can be established.

The advantages of the wire aligner lie in its lightweight design and low friction. 
In conventional coupled tendon-driven systems, the required number and weight of pulleys increase proportionally with the number of joints they traverse. 
In contrast, this method enables power transmission between the two links without passing through intermediate joints by using only one set of wire aligners. 
Furthermore, this approach employs grooved pulleys with built-in bearings to guide the wire, eliminating sliding parts and reducing friction.

A similar approach involves placing a wire alignment mechanism on one of the links and anchoring the wire end to the other link \cite{ZHANG2020103693}. 
In this method, power transmission via the wire is limited to specific two links. 
In contrast, our approach uses pairs of wire alignment mechanisms, allowing power transmission to links other than the two links equipped with the wire alignment mechanism. 
Therefore, our method can transmit power to links further away than existing methods.
}%
{%


受動3次元ワイヤ整列装置の詳細な構造を\figref{figure:wirealignerstructure}に示す.
ワイヤがこの装置を経由すると, ワイヤが装置に対してどのような角度で入射しても既定の直線に一致する.
2つのリンクにワイヤ整列装置をそれぞれ設置することで, 2リンク間を直線で結ぶワイヤ経路を作成できる.

ワイヤ整列装置のメリットとしては, 軽量であることと摩擦が小さいことの2点がある.
一般的な干渉ワイヤ駆動では経由する関節数に比例して必要なプーリの数・重量が増加してしまう.
一方で本手法ではワイヤ整列装置1組のみを用いることで途中の関節を経由することなく, 2リンク間で動力を伝達することが可能である.
さらに本手法ではベアリングを内蔵した溝つきプーリによってワイヤを誘導するため, 摺動部が無く摩擦を低減できる.





類似の手法としては, ワイヤ整列機構を片方のリンクに一つ配置しもう片方のリンクにワイヤ端を固定した例が存在する\cite{ZHANG2020103693}.
この手法では特定の2リンク間でのみワイヤによる動力伝達が行われる.
一方で本手法ではワイヤ整列機構をペアで使用するため, ワイヤ整列機構を有する2リンク以外のリンクにも動力を伝達することが可能である.
このため本手法は既存手法よりも遠くのリンクまで動力を伝達可能である.












}%

\begin{figure*}[t]
  \centering
  \includegraphics[width=2.0\columnwidth]{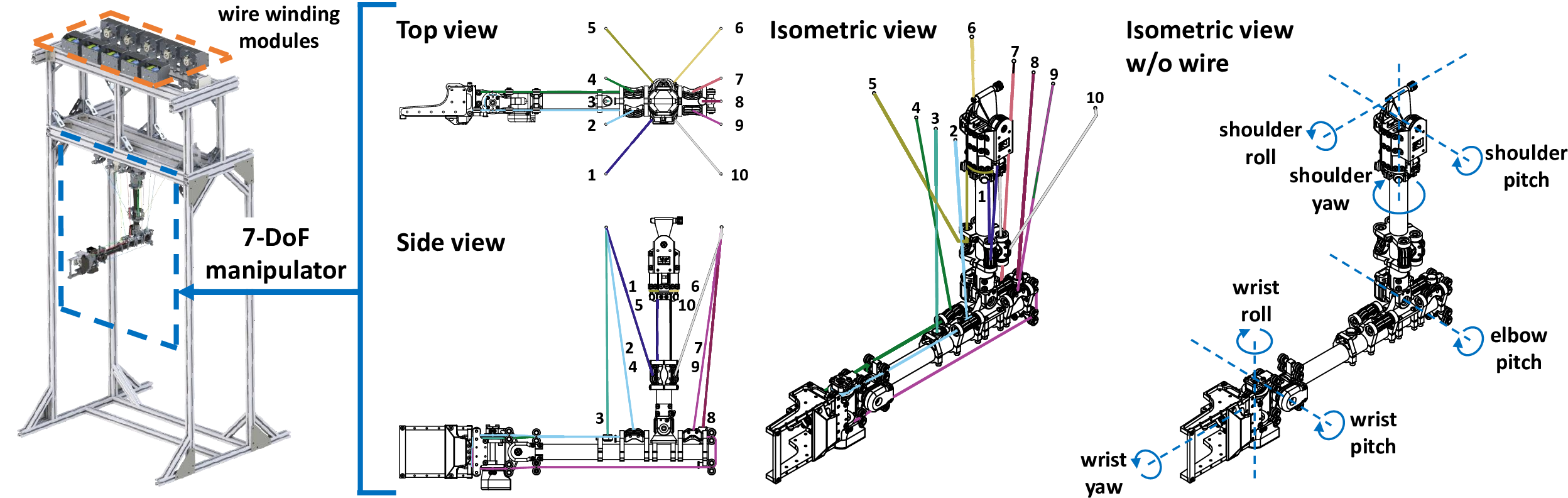}
  \vspace{-1ex}
  \caption{Overview of SAQIEL.}
  \label{figure:gl80armoverview}
  \vspace{-1.5ex}
\end{figure*}

\subsection{Design of Passive 3D Wire Aligner} \label{subsec:aligner-design}
\switchlanguage%
{%
  As shown in \figref{figure:wirealignerstructure}, the passive 3D wire aligner automatically accommodates the wire's pitch and roll displacements upon entry. 
  The alignment in the pitch direction of the wire is achieved by winding the wire around the grooved pulley. 
  This grooved pulley incorporates needle bearings, enabling low-friction wire transmission. 
  Additionally, by sandwiching the wire between two grooved pulleys, the wire is prevented from dislodging from the aligner.

  Alignment in the roll direction of the wire is achieved through rotation of the pulley holder. 
  This rotational degree of freedom is facilitated by deep groove ball bearings attached at both ends of the pulley holder.

  In this study, a passive 3D wire aligner was designed for use with a Vectran rope of diameter \SI{1}{\milli\metre} (tensile strength of \SI{1560}{\newton}, VB-175, Hayami Industry), assuming a maximum tension of \SI{500}{\newton}. 
  By adopting the slim \SI{1}{\milli\metre} diameter wire, the device was successfully miniaturized and lightweighted to dimensions of \SI{44}{\milli\metre} $\times$ \SI{16}{\milli\metre} $\times$ \SI{21}{\milli\metre} and a weight of \SI{27}{\gram}.
}%
{
\figref{figure:wirealignerstructure}に示すとおり, 受動3次元ワイヤ整列装置はワイヤ入射角のピッチ方向変位とロール方向変位に自動的に追従する.
ワイヤのピッチ方向変位への追従は溝つきプーリにワイヤを巻きつかせることで実現している.
この溝つきプーリにはニードルベアリングが内蔵されており低摩擦なワイヤの伝達が可能である.
また溝つきプーリ２つの間にワイヤを挟むことによってワイヤが整列装置から外れることを防止している.

ワイヤのロール方向への追従は溝つきプーリのホルダが回転することで実現している.
プーリホルダの両端に取り付けられた深溝玉軸受によってこの回転自由度が生み出されている.

本研究では直径\SI{1}{\milli.\metre}のVectran rope (tensile strength of \SI{1560}{\newton}, VB-175, Hayami industry) を最大張力\SI{500}{\newton}で使用する想定で受動3次元ワイヤ整列装置を設計した.
直径\SI{1}{\milli.\metre}の細径ワイヤを採用することで, 寸法\SI{44}{\milli\metre}$\times$\SI{16}{\milli\metre}$\times$\SI{21}{\milli\metre}, 重量\SI{27}{\gram} と装置全体の小型軽量化に成功した.

}





\section{Design of 7-DoF manipulator with Passive 3D Wire Aligner} \label{sec:design}
\subsection{Overview} \label{subsec:design-overview}
\switchlanguage%
{%
In this chapter, we provide a detailed description of the design of the 7-degree-of-freedom manipulator, termed SAQIEL ({\bf SA}fe, {\bf Q}u{\bf I}ck and {\bf E}xtremely {\bf L}ightweight manipulator), created for the purpose of validating the performance of the passive 3D wire aligner.

As mentioned in \secref{sec:introduction}, reducing the effective mass of the manipulator is crucial to enhance safety during collisions. 
Moreover, improving joint backdrivability is effective for enhancing safety during quasi-static contact scenarios such as crushing or clamping. 
Therefore, the design objectives of SAQIEL are as follows:
\begin{enumerate}
  \item Minimization of effective mass.
  \item Reduction of friction in the power transmission system.
\end{enumerate}
Furthermore, these design objectives should be achieved while ensuring payload capacity, accuracy, and operational speed.

For design goal 1) , SAQIEL has all drive motors located on the root link. 
An overview of SAQIEL is presented in \figref{figure:gl80armoverview}. 
The structure involves 10 motors on the root link, each winding wires to control the 7-DoF manipulator.
Power transmission from the root link to each link is facilitated by 10 sets of passive 3D wire aligners.

For design goal 2), SAQIEL employs a mechanism in which motors directly wind the wires without the use of gear or belt-based reduction mechanisms. 
Detailed specifications of the gearless wire modules are explained in \secref{subsec:design-wire-module}. 
Additionally, through the use of multiple small pulleys in the wire path to eliminate sliding parts, low-friction power transmission is achieved.

The primary specifications of SAQIEL are presented in \tabref{table:system_params}. 
SAQIEL's arm length is \SI{0.78}{\metre}, similar to a human arm. 
However, the total weight of the moving parts (all links except the root link) is approximately \SI{1.5}{\kilogram}. 
This is about \SI{60}{\percent} lighter compared to the weight of a human upper limb of the same length (\SI{4.1}{\kilogram}) \cite{latella2019human-gazebo}.

SAQIEL features two types of wire paths: linear and circular. 
The definitions of each path are as follows:
\begin{itemize}
  \item Linear wire path: A straight line connecting specific points on two links.
  \item Circular wire path: An arc coaxial with the joint axis.
\end{itemize}
Each path has its respective advantages: linear paths facilitate larger moment arms, while circular paths enable larger range of motion. 
Balancing lightweight design and range of motion, the choice of wire paths is made for each joint.
}%
{%
本章では受動3次元ワイヤ整列装置の性能検証のために作成した7自由度マニュピレータSAQIEL ({\bf SA}fe , {\bf Q}u{\bf I}ck and {\bf E}xtremely {\bf L}ightweight manipulator)の設計について詳細な解説を行う.

\secref{sec:introduction}で述べたとおり, 衝突時の安全性を高めるためにはマニュピレータの有効質量を小さくすることが重要である.
さらに準静的な接触(押しつぶしやクランピング)における安全性の向上には, 関節のバックドライバビリティを高めることが有効である.
そこでSAQIELの設計目標を以下の2つとする.
\begin{enumerate}
  \item 有効質量の最小化
  \item 駆動系の摩擦の低減
\end{enumerate}
さらにこれらの設計目標を損なわない範囲で, ペイロード・正確性・動作速度も確保する.



設計目標1(有効質量の最小化)のために, SAQIELでは全ての駆動用モータをルートリンクに搭載した.
\figref{figure:gl80armoverview}にSAQIELの概要を示す.
ルートリンク上の10個のモータがそれぞれワイヤを巻き取ることで7自由度の関節を制御する構造となっている.
ルートリンクから各リンクへの動力伝達は10組の受動3次元ワイヤ整列装置によって行っている.

設計目標2(駆動系の摩擦の低減)のために, SAQIELではギアやベルトによる減速機構を使用せずモータでワイヤを直接巻き取る機構を採用した.
ギアレスワイヤモジュールの詳細な仕様については\secref{subsec:design-wire-module}で解説する.
さらにワイヤ経路において小型プーリを多用し摺動部を排除することで, 低摩擦な動力伝達を実現した.

\tabref{table:system_params}にSAQIELの主要なスペックを示す.
SAQIELの上肢長は\SI{0.78}{\metre}と人間の腕部に近い寸法となっている.
一方で動作部(ルートリンク以外の全リンク)の合計重量は約\SI{1.5}{\kilogram}となっている.
これは同じ長さ(\SI{0.78}{\metre})の人間の上肢の重量(\SI{4.1}{\kilogram})と比べて\SI{60}{\percent}ほど軽量である\cite{latella2019human-gazebo}.

SAQIELのワイヤの経路には直線型と円型の2種類が存在する.
各経路の定義を以下に示す.
\begin {itemize}
  \item 直線型ワイヤ経路: 特定の2リンク上の点を結ぶ直線 
  \item 円型ワイヤ経路: 関節軸と同軸な円弧
\end {itemize}
直線型経路はモーメントアームを大きくしやすく, 円形経路は可動域を大きくしやすいというメリットがそれぞれ存在する.
軽量性や可動範囲の兼ね合いから, 各関節におけるワイヤ経路を選択する.

}%

\begin{figure}[t]
  \centering
  \includegraphics[width=1.0\columnwidth]{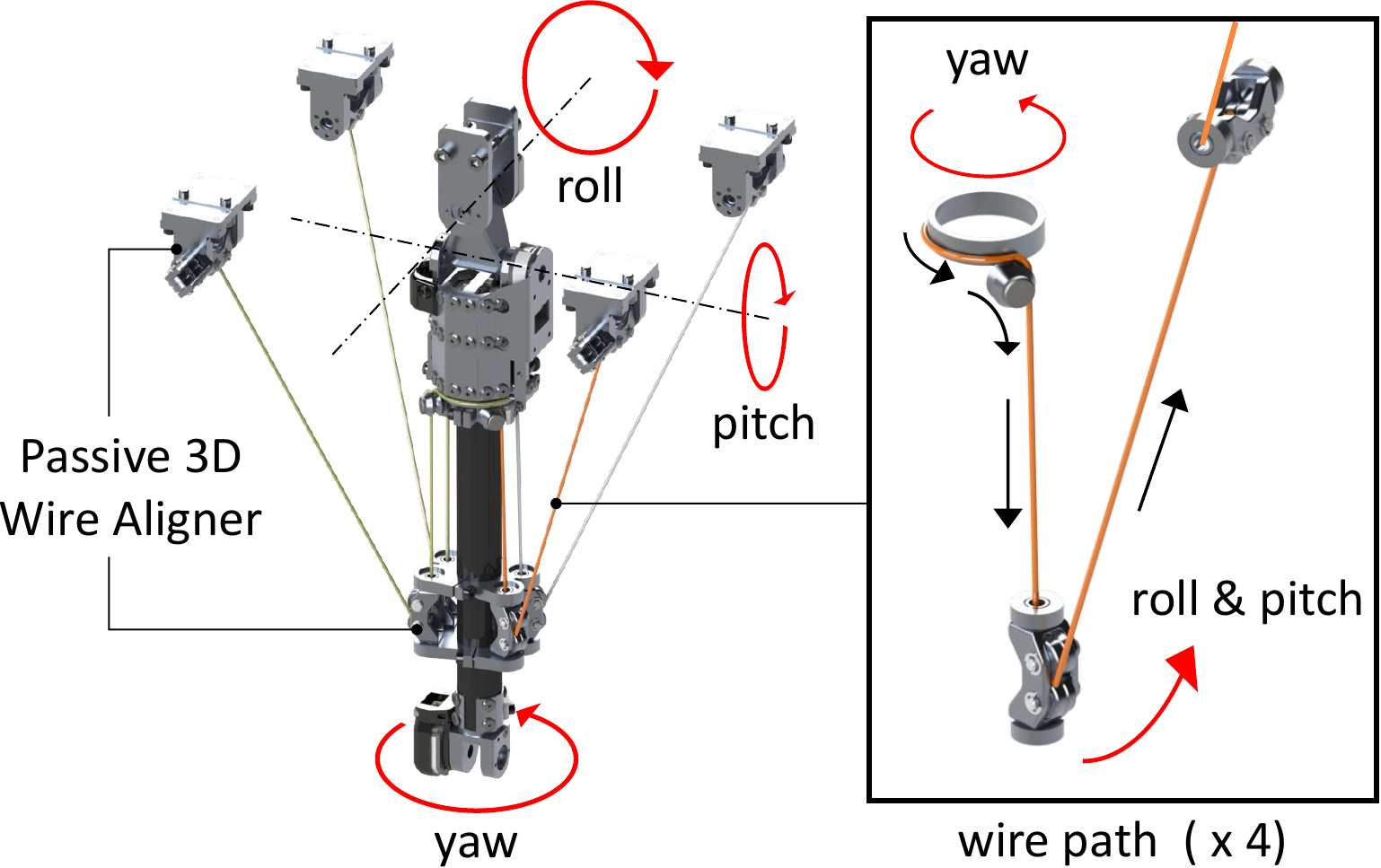}
  \vspace{-2ex}
  \caption{Shoulder design of SAQIEL.}
  \label{figure:gl80armshoulder}
  \vspace{-1ex}
\end{figure}

\begin{figure}[t]
  \centering
  \includegraphics[width=1.0\columnwidth]{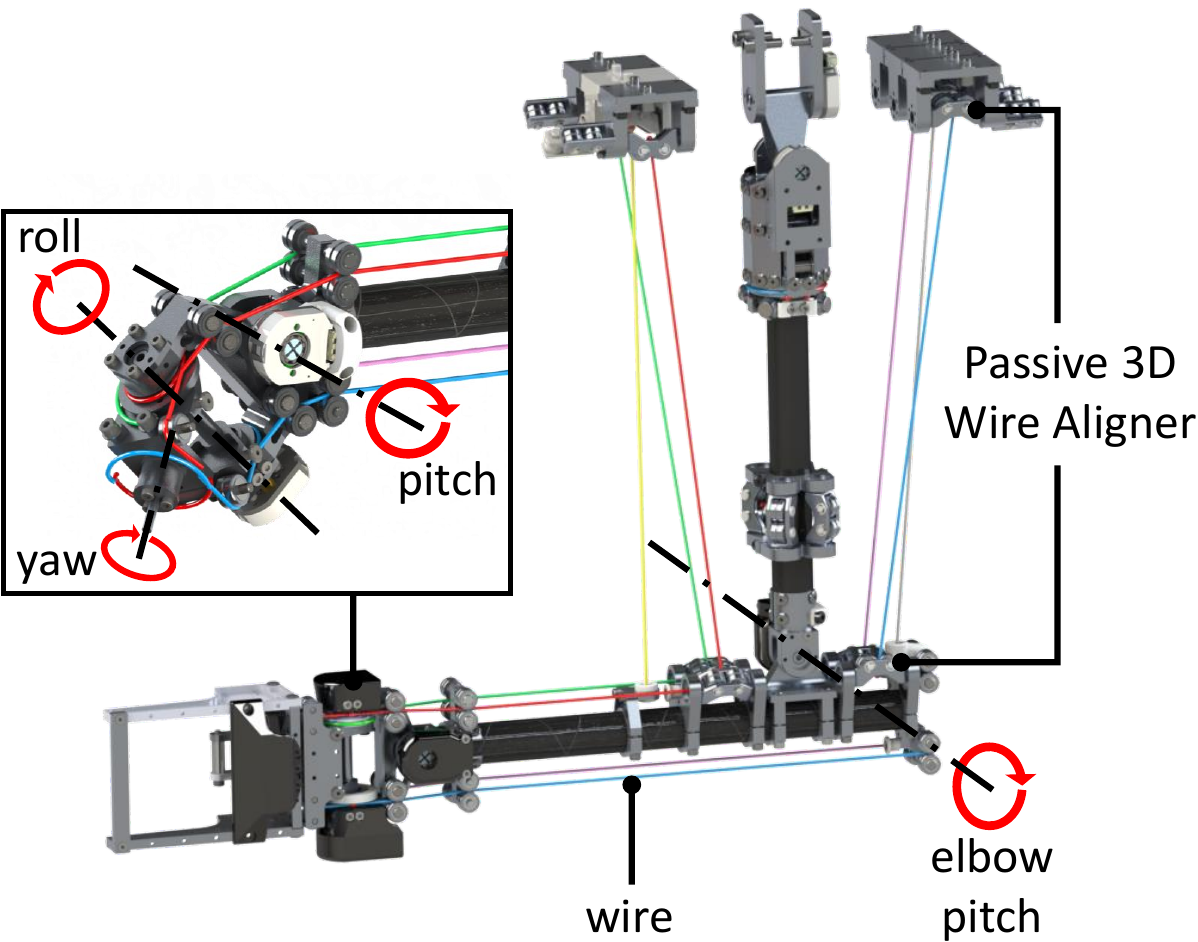}
  \vspace{-2ex}
  \caption{Elbow and wrist design of SAQIEL.}
  \label{figure:gl80armelbowwrist}
  \vspace{-1ex}
\end{figure}

\subsection{Shoulder Mechanism} \label{subsec:design-shoulder}
\switchlanguage%
{%
In the design of the manipulator's shoulder joint, it is crucial to balance sufficient joint torque and lightweight characteristics. 
Conventional coupled tendon-driven robots \cite{hirose1991tendon, yokoi1991design} employ numerous large-diameter pulleys on the shoulder section to ensure joint torque. 
However, the extensive use of large-diameter pulleys complicates the mechanism and increases its weight. 
In the shoulder component of SAQIEL, a passive 3D wire aligner is employed to achieve balance between a large moment arm and lightweight design.

Detailed structures of the three shoulder joints (roll, pitch, yaw) are depicted in \figref{figure:gl80armshoulder}. 
The shoulder roll, pitch, and yaw joints are mainly controlled by four wires (wires 1, 5, 6, 10) arranged circularly around the upper arm's long axis. 
These four wires follow the path: root link $\to$ upper arm link $\to$ shoulder link.

Four sets of passive 3D wire aligners are placed on the root link and the upper arm link, connecting the two links with a linear wire path. 
This configuration allows for a significant moment arm (approximately \SI{90}{\milli\metre}) around the shoulder roll and pitch axes. 
Furthermore, due to the wires not passing through the interior of the shoulder joint, a lightweight shoulder joint (weighing around \SI{0.5}{\kilogram}, including the shoulder joint and upper arm link) with fewer large-diameter pulleys can be realized.

The upper arm link to the shoulder link is connected by a circular wire path around the shoulder yaw axis. 
The four wires passing through the passive 3D wire aligner on the upper arm link wind around coaxial pulleys on the shoulder link. 
The ends of these wires wound around the pulley are fixed on the shoulder link.
}%
{%
  マニュピレータの肩関節の設計において重要な点は十分な関節トルクと軽量性を両立することである.
  一般的な干渉ワイヤ駆動ロボット\cite{hirose1991tendon, yokoi1991design}では, 関節トルクを確保するために肩部に多数の大径プーリを搭載している.
  一方で大径プーリの多用は機構の複雑化と重量増加を招く.
  SAQIELの肩パーツでは, 受動3次元ワイヤ整列装置を用いて大モーメントアームと軽量性を両立した.

  \figref{figure:gl80armshoulder}に肩3関節(shoulder roll, shoulder pitch, shoulder yaw)の詳細な構造を示す.
  肩3関節は上腕長軸周りに環状に配置された4本のワイヤ(ワイヤ1, 5, 6, 10)によって主に制御されている.
  これら4本のワイヤはルートリンク-上腕リンク-肩リンクという経路を辿る.

  ルートリンクと上腕リンクには4組の受動3次元ワイヤ整列装置が設置されており, 2リンク間は直線型ワイヤ経路で結ばれている.
  これによってshoulder roll軸とshoulder pitch軸回りに大きなモーメントアーム(約\SI{90}{\milli \metre})を確保した.
  さらに本構造ではワイヤが肩関節内部を経由しないため, 大径プーリの少ない軽量な肩関節(肩関節と上腕リンク合わせて\SI{0.5}{\kilogram})を実現できる.

  上腕リンク-肩リンク間はshoulder yaw軸周りの円型ワイヤ経路で結ばれている.
  4本のワイヤは上腕リンクの受動3次元ワイヤ整列装置を通り, 肩リンク上のyaw軸に同軸なプーリに巻き付く.
  このプーリに巻き付いたワイヤの端部は肩リンク上に固定される.









}

\subsection{Elbow and Wrist Mechanism} \label{subsec:design-elbow-wrist}
\switchlanguage%
{%

To reduce the effective mass of the manipulator, lightweight design of links near the end effector is crucial. 
Therefore, in the design of SAQIEL's elbow pitch joint and three joints of wrist (pitch, roll, yaw), the goal was to create a lightweight configuration with the minimal required number of parts.

Detailed designs of the elbow pitch joint and three joints of wrist are presented in \figref{figure:gl80armelbowwrist}. 
The elbow pitch joint is controlled by six linear wire paths (wires 2, 3, 4, 7, 8, 9) connecting the root link and the forearm link. 
These six wire paths coincide with the straight lines connecting six passive 3D wire aligners on the root link side as well as four passive 3D wire aligners and two wire termination components on the forearm link side. 
By adopting linear wire paths similar to the shoulder roll and pitch joints, balance between lightweight design (\SI{0.5}{\kilogram}) and a large moment arm is achieved.

Among the wires driving the elbow pitch joint, four (wires 2, 4, 7, 9) extend from the forearm link to the end effector link, controlling the wrist's three joints (wrist pitch, wrist roll, wrist yaw). 
The range of motion of the wrist joints is based on the human body. 
A linear wire path is used to drive the wrist pitch joint, which has a small range of motion, and a circular wire path is used to drive the wrist roll and yaw joints, which have a large range of motion.
Moreover, considering that the required torque for the wrist joints is smaller compared to the shoulder and elbow, a configuration using small pulleys was employed for creating the wire paths instead of 3D passive wire aligners. 
By choosing appropriate wire paths and path creation methods based on the required joint torques and ranges of motion, a lightweight (\SI{0.5}{\kilogram}) and large-range-of-motion wrist joint was realized.
}%
{%
マニュピレータの有効質量を小さくするためには, エンドエフェクタに近いリンクの軽量化が非常に重要である.
そのためSAQIELのelbow pitch関節と手首3関節(wrist pitch, wrist roll, wrist yaw)の設計では, 必要最小限のパーツ数で軽量に構成することを目標とした.

\figref{figure:gl80armelbowwrist}にelbow pitch関節と手首3関節の詳細な設計を示す.
elbow pitch関節はルートリンク-前腕リンク間を結ぶ6本の直線型ワイヤ経路(ワイヤ2, 3, 4, 7, 8, 9)によって制御されている.
これら6本のワイヤ経路は, ルートリンク側の6個の受動3次元ワイヤ整列装置と, 前腕リンク側の4個の受動3次元ワイヤ整列装置+2個のワイヤ端固定部品間を結ぶ直線と一致する.
shoulder roll関節やshoulder pitch関節と同様に直線型ワイヤ経路を採用することで, 軽量(\SI{0.5}{\kilogram})かつ大モーメントアームな肘関節を実現した.

elbow pitch関節を駆動するワイヤのうち4本(ワイヤ2, 4, 7, 9)は前腕リンクから手先リンクまで伸び, 手首3関節(wrist pitch, wrist roll, wrist yaw)を制御している.
手首関節の可動域は人体を参考にしており, 可動域の小さいwrist pitch関節の駆動には直線型ワイヤ経路を, 可動域の大きいwrist roll関節とwrist yaw関節の駆動には円形ワイヤ経路をそれぞれ採用している.
また手首3関節は必要トルクが肩や肘に比べて小さいことから, 3次元受動ワイヤ整列装置ではなく小型プーリのみを使用してワイヤ経路を作成している.
必要関節トルクや可動域に応じて適切なワイヤ経路や経路作成方法を選択することで, 軽量(\SI{0.5}{\kilogram})かつ可動域の大きい手首関節を実現した.










}

\begin{table}[t]
	\centering
	\caption{Physical parameters of SAQIEL.}
	\label{table:system_params}
	\begin{tabular}{crc}
		\toprule
		\multicolumn{2}{c}{Items}  						     		& Value \\
		\midrule
		\multirow{4}{6em}{Mass without electrical wiring} 
											& total moving part     & \SI{1.5}{\kilogram} \\
											& upper arm        		& \SI{0.5}{\kilogram} \\
											& forearm        		& \SI{0.5}{\kilogram} \\
											& wrist        			& \SI{0.5}{\kilogram} \\
		\midrule
		\multirow{7}{6em}{Range of motion} 	
											& shoulder roll    		&  \SI{-55}{\deg}$\sim$\SI{55}{\deg} \\
											& shoulder pitch   		&  \SI{-55}{\deg}$\sim$\SI{55}{\deg} \\
											& shoulder yaw   		&  \SI{-90}{\deg}$\sim$\SI{90}{\deg} \\
											& elbow			   		&  \SI{-60}{\deg}$\sim$\SI{60}{\deg} \\
											& wrist pitch   		&  \SI{-45}{\deg}$\sim$\SI{45}{\deg} \\
											& wrist roll   			&  \SI{-80}{\deg}$\sim$\SI{80}{\deg} \\
											& wrist yaw   			&  \SI{-150}{\deg}$\sim$\SI{150}{\deg} \\
		\midrule
		\multirow{3}{6em}{Distance between joints} 	
											& shoulder roll - elbow pitch   &  \SI{0.34}{\metre} \\
											& elbow pitch - wrist pitch		&  \SI{0.24}{\metre} \\
											& wrist pitch - fingertip	&  \SI{0.20}{\metre} \\
    \midrule
    \multicolumn{2}{l}{Upper limb length}   					& \SI{0.78}{\metre} \\
		\midrule
		\multicolumn{2}{l}{Maximum wire tension}   					& \SI{490}{\newton} \\
		\midrule
		\multicolumn{2}{l}{Maximum wire speed (no load)} 			& \SI{0.86}{\metre/\second} \\
		\bottomrule
	\end{tabular}
  \vspace{-1ex}
\end{table}

\subsection{Gear-less Wire Winding Module Mechanism} \label{subsec:design-wire-module}
\begin{figure}[t]
  \centering
  \includegraphics[width=1.0\columnwidth]{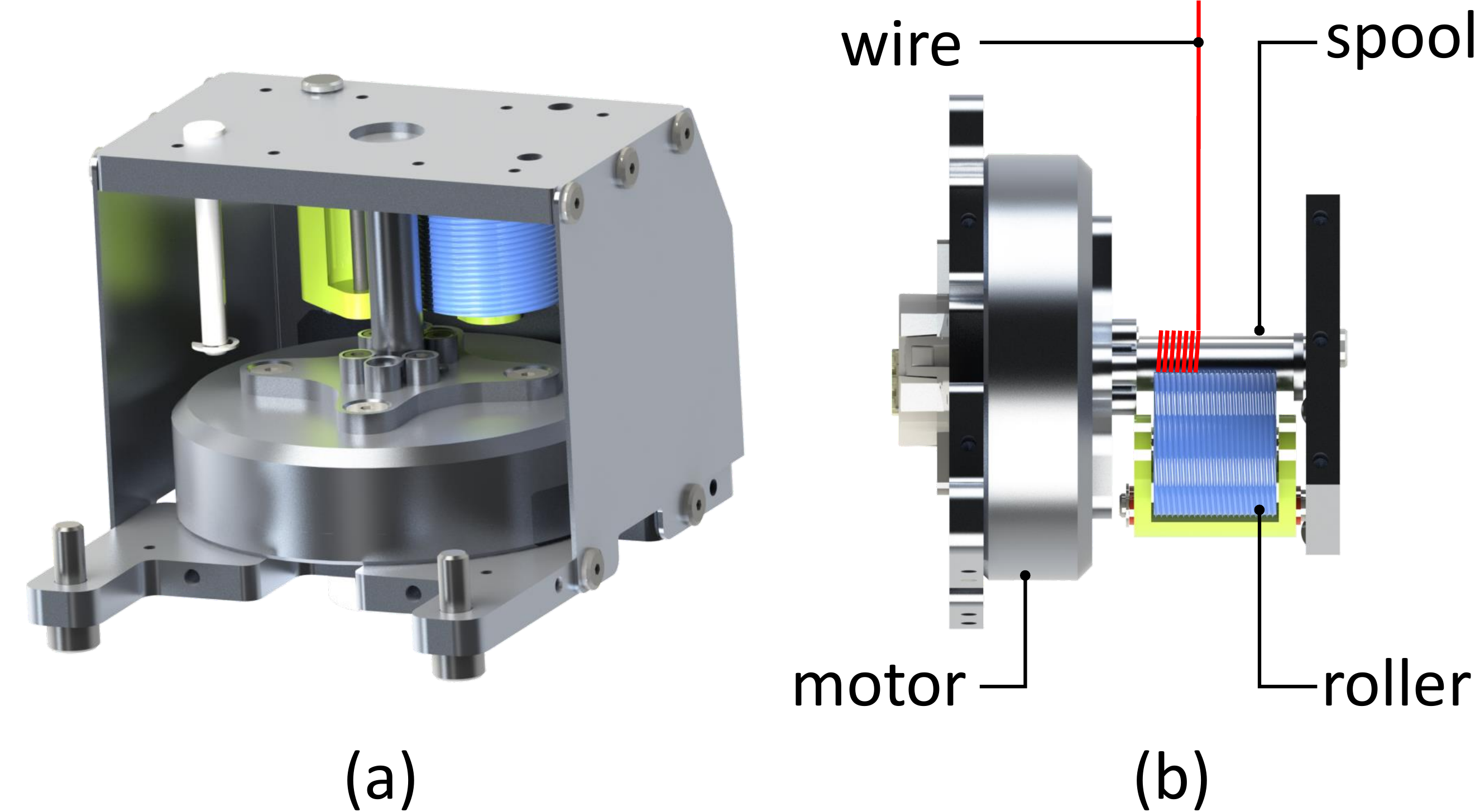}
  \vspace{-2ex}
  \caption{Wire winding module design of SAQIEL. (a) isometoric view of the module, (b) winding mechanism of the module.}
  \label{figure:wirewindingmodule}
  \vspace{-0.5ex}
\end{figure}

\switchlanguage%
{%
As mentioned in \secref{subsec:design-overview}, the design requirement of this mechanism is to achieve low-rcition wire winding without using a gearbox. 
In this study, a high torque constant BLDC gimbal motor (T-MOTOR GL80 KV30) was utilized to achieve low-friction wire winding without the need for a gearbox.

Detailed design of the gearless wire module is presented in \figref{figure:wirewindingmodule}. 
\SI{1}{\milli.\metre} diameter Vectran rope (tensile strength of \SI{1560}{\newton}, VB-175, Hayami industry) is wound around a \SI{10}{\milli.\metre} diameter pulley fixed to the motor. 
The peak tension is \SI{490}{\newton}, providing sufficient tension to drive each joint.

To prevent wire entanglement and ensure accurate winding length, a grooved roller is pressed against the wire on the pulley's surface. 
This design feature effectively prevents wire entanglement and ensures precise winding length.
}%
{%
\secref{subsec:design-overview}で述べたとおり,本機構の設計要件は減速機がなく摩擦の小さいワイヤ巻取りを実現することである.
本研究ではトルク定数の大きいBLDCジンバルモータ(T-MOTOR GL80 KV30)を使用することで, 減速機が無く低摩擦なワイヤ巻取りを実現した.

\figref{figure:wirewindingmodule}にギヤレスワイヤモジュールの詳細な設計を示す.
直径\SI{1}{\milli.\metre}のVectran rope (tensile strength of \SI{1560}{\newton}, VB-175, Hayami industry)を, モータに固定された直径\SI{10}{\milli.\metre}のプーリで巻き取っている.
ピーク張力は\SI{490}{\newton}となっており, 各関節を駆動するのに十分な張力を発揮可能である.

プーリ上のワイヤに溝つきのrollerを押し当てることでワイヤの絡まりを防止し, 正確な巻取り長さを実現している.

}

\begin{figure*}[t]
  \centering
  \includegraphics[width=2.0\columnwidth]{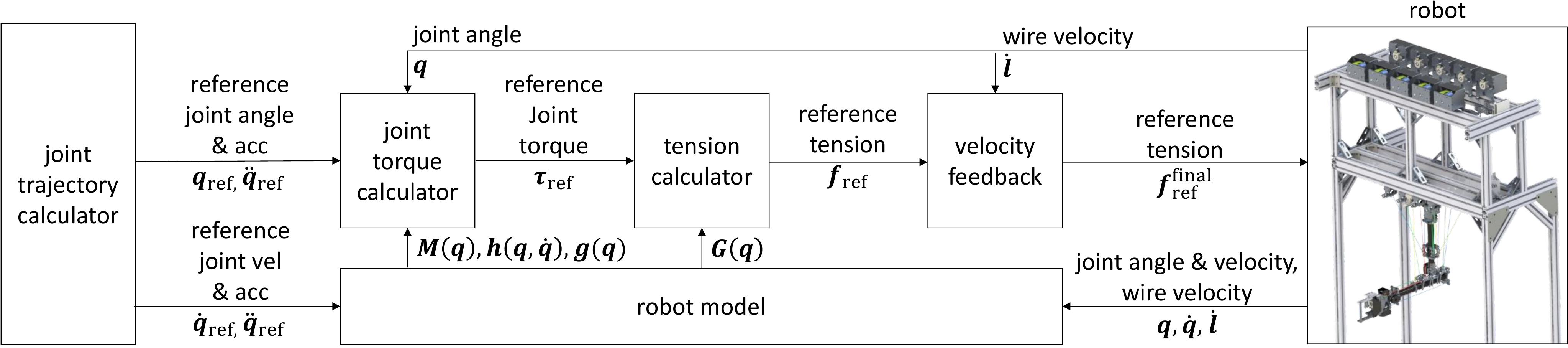}
  \vspace{-1ex}
  \caption{Controller of SAQIEL.}
  \label{figure:controller}
  \vspace{-1.5ex}
\end{figure*}

\section{Controller of 7-DoF manipulator with Passive 3D Wire Aligner} \label{sec:controller}
\switchlanguage%
{%
The controller of SAQIEL is depicted in \figref{figure:controller}. 
SAQIEL employs a control strategy based on the Computed Torque Method. 
However, due to the adoption of the coupled tendon-driven mechanism, the desired joint torques $\bm{\tau}_{\mathrm{ref}}$ are transformed into target wire tensions $\bm{f}_{\mathrm{ref}}$ after computation \cite{kawamura2016jointspace}.

The torques $\bm{\tau}$ at the joints generated by the tension $\bm{f}$ of the wires can be calculated using the following equation:
\begin{align}
  \label{eq:tension-torque}
  \bm{\tau} = -G^{\mathrm{T}} \bm{f}
\end{align}
here, $G$ is a matrix representing the moment arm of each wire at each joint, known as the muscle Jacobian. 
The muscle Jacobian $G(\bm{q})$ at a specific joint angle $\bm{q}$ is defined as:
\begin{align}
  \label{eq:muscle-jacobian}
  G(\bm{q}) = \frac{\partial \bm{l}}{\partial \bm{q}}
\end{align}
where $\bm{l}$ is the current wire length. 
Using Equation \eqref{eq:tension-torque}, the target wire tension $\bm{f}_{\mathrm{ref}}$ that satisfies the target joint torque $\bm{\tau}_{\mathrm{ref}}$ is determined by solving the following quadratic programming problem \cite{kawamura2016jointspace}:
\begin{align}
  \label{calc-tension}
  \begin{split}
    \text{minimize} & \quad |\bm{f_{\mathrm{ref}}}|^{2}+(\bm{\tau}_{\mathrm{ref}}+G^{T}\bm{f}_{\mathrm{ref}})^{T} \Lambda (\bm{\tau}_{\mathrm{ref}}+G^{T}\bm{f}_{\mathrm{ref}})
    \\
    \text{subject to} & \qquad \qquad \bm{f}_{\mathrm{min}} \leq \bm{f}_{\mathrm{ref}} \leq \bm{f}_{\mathrm{max}}
\end{split}
\end{align}
here, $\Lambda$ is the weight matrix for each joint torque, $\bm{f}_{\mathrm{min}}$ / $\bm{f}_{\mathrm{max}}$ is the minimum / maximum tension for each wire. 
The first term of the objective function minimizes the magnitude of wire tension, while the second term minimizes the error between the target joint torque and the actual joint torque generated by wire tension. 
Solving this quadratic programming problem allows us to determine the minimized target wire tension $\bm{f}_{\mathrm{ref}}$ while satisfying the target joint torque.

Since the moment arm of linear wire paths varies with joint angles, the muscle Jacobian $G(\bm{q})$ needs to be recalculated for each change in joint angle. 
A muscle Jacobian calculation library that accommodates both linear and circular wire paths was developed in this study. 
This library enables the conversion from desired joint torques to target wire tensions to be performed at a high frequency of over \SI{500}{\hertz}.

The calculation of the desired joint torques $\bm{\tau}_{\mathrm{ref}}$ is derived from the following equation:
\begin{align}
  \label{eq:joint-torque}
  \bm{\tau}_{\mathrm{ref}} = \bm{M}(\bm{q}) (\bm{K}_{\mathrm{p}} (\bm{q}_{\mathrm{ref}} - \bm{q}) + \bm{\ddot{q}}_{\mathrm{ref}}) + \bm{h}(\bm{q}, \bm{\dot{q}}) + \bm{g}(\bm{q})
\end{align}
where $\bm{q}$ represents the current joint angle vector, $\bm{M}(\bm{q})$ is the inertia matrix, $\bm{K}_{\mathrm{p}}$ is the position feedback gain matrix, $\bm{q}_{\mathrm{ref}}$ is the target joint angle, $\bm{\ddot{q}}_{\mathrm{ref}}$ is the target joint acceleration derived from the desired trajectory, $\bm{h}(\bm{q}, \bm{\dot{q}})$ is the vector representing centrifugal and Coriolis forces, and $\bm{g}(\bm{q})$ is the gravity vector. 

Notably, unlike the conventional computed torque method, the feedback term concerning joint angular velocity $\bm{\dot{q}}$ is absent. 
This is due to the use of wire velocity $\bm{\dot{l}}$ rather than joint angular velocity $\bm{\dot{q}}$ for velocity feedback. 
Delay arises between the motor output shaft angular velocity and the joint angular velocity due to the elasticity of the wire used for power transmission. 
Consequently, using feedback based on joint angular velocity $\bm{\dot{q}}$ (acquired from joint encoders) could lead to manipulator oscillations. 
Therefore, this study employs feedback based on wire velocity $\bm{\dot{l}}$ (acquired from encoders attached to the motors). 
The final target wire tensions $\bm{f}_{\mathrm{ref}}^{\mathrm{final}}$ including the velocity feedback term are calculated using the following equation:
\begin{align}
  \label{eq:wire-tension}
  \bm{f}_{\mathrm{ref}}^{\mathrm{final}} = \bm{K}_{\mathrm{v}} (\bm{\dot{l}}_{\mathrm{ref}} - \bm{\dot{l}}) + \bm{f}_{\mathrm{ref}}
\end{align}
here, $\bm{K}_{\mathrm{v}}$ is the velocity feedback gain matrix, and $\bm{\dot{l}}_{\mathrm{ref}}$ represents the desired wire velocity.
}%
{%

SAQIELの制御器を\figref{figure:controller}に示す.
SAQIELは計算トルク法を基本とした制御器を用いている.
ただしワイヤ干渉駆動を採用しているため, 目標関節トルク$\bm{\tau}_{\mathrm{ref}}$を求めた後に目標ワイヤ張力$\bm{f}_{\mathrm{ref}}$へ変換している.

ワイヤ張力$\bm{f}$が関節に発生させるトルク$\bm{\tau}$は以下の式で求められる.
\begin{align}
  \label{eq:tension-torque}
  \bm{\tau} = -G^{\mathrm{T}} \bm{f}
\end{align}
ただし$G$は各関節における各ワイヤのモーメントアームを表す行列であり, 筋長ヤコビアンと呼ばれる.
ある関節角度$\bm{q}$における筋長ヤコビアン$G(\bm{q})$は以下の式で定義される.
\begin{align}
  \label{eq:muscle-jacobian}
  G(\bm{q}) = \frac{\partial \bm{l}}{\partial \bm{q}}
\end{align}
ただし$\bm{l}$は現在のワイヤ長さである.
式\eqref{eq:tension-torque}を用いて, 目標関節トルク$\bm{\tau}_{\mathrm{ref}}$を満たす目標ワイヤ張力$\bm{f}_{\mathrm{ref}}$を以下の二次計画問題を解いて求める\cite{kawamura2016jointspace}.
\begin{align}
  \label{calc-tension}
  \begin{split}
    \text{minimize} & \quad |\bm{f_{\mathrm{ref}}}|^{2}+(\bm{\tau}_{\mathrm{ref}}+G^{T}\bm{f}_{\mathrm{ref}})^{T} \Lambda (\bm{\tau}_{\mathrm{ref}}+G^{T}\bm{f}_{\mathrm{ref}})
    \\
    \text{subject to} & \qquad \qquad \bm{f}_{\mathrm{min}} \leq \bm{f}_{\mathrm{ref}} \leq \bm{f}_{\mathrm{max}}
\end{split}
\end{align}
ただし, $\Lambda$は各関節トルクの重み行列, $\bm{f}_{\mathrm{min}}$ / $\bm{f}_{\mathrm{max}}$ は各ワイヤの最小/最大張力である.
目的関数の第一項はワイヤ張力の大きさを最小化する項, 第二項は目標関節トルクとワイヤ張力が生み出す実際の関節トルクの誤差を最小化する項である.
この二次計画法を解くことで, 目標関節トルクを満たしつつワイヤ張力を最小化する目標ワイヤ張力$\bm{f}_{\mathrm{ref}}$を求めることができる.

直線型ワイヤ経路のモーメントアームは関節角度に応じて変化するため, 筋長ヤコビアン$G(\bm{q})$も関節角度$\bm{q}$の変化に応じて都度計算する必要がある.
本研究では直線型ワイヤ経路と円型ワイヤ経路の両者に対応可能な筋長ヤコビアン計算ライブラリを作成した.
本ライブラリによって目標関節トルクから目標ワイヤ張力への変換を\SI{500}{\hertz}以上の高周期で行うことができる.

目標関節トルク$\bm{\tau}_{\mathrm{ref}}$は以下の式から計算される.
\begin{align}
  \label{eq:joint-torque}
  \bm{\tau}_{\mathrm{ref}} = \bm{M}(\bm{q}) (\bm{K}_{\mathrm{p}} (\bm{q}_{\mathrm{ref}} - \bm{q}) + \bm{\ddot{q}}_{\mathrm{ref}}) + \bm{h}(\bm{q}, \bm{\dot{q}}) + \bm{g}(\bm{q})
\end{align}
ただし$\bm{M}(\bm{q})$は慣性行列, $\bm{K}_{\mathrm{p}}$は位置のフィードバックゲイン行列, $\bm{q}_{\mathrm{ref}}$は目標関節角度, $\bm{\ddot{q}}_{\mathrm{ref}}$は目標軌道から定まる目標関節角加速度, $\bm{h}(\bm{q}, \bm{\dot{q}})$は遠心力およびコリオリ力を表すベクトル, $\bm{g}(\bm{q})$は重力ベクトルである.
上式には通常の計算トルク法と異なり, 関節角速度$\bm{\dot{q}}$についてのフィードバック項が含まれていない.
これは, 速度についてのフィードバックを関節角速度$\bm{\dot{q}}$ではなくワイヤ速度$\bm{\dot{l}}$を用いて行うためである.
動力伝達に用いるワイヤには弾性があるため, モータ出力軸の角速度と関節の角速度の間には遅延が発生する.
そのため関節角速度$\bm{\dot{q}}$(関節エンコーダから取得)に対するフィードバックを行うとマニュピレータが発振する恐れがある.
そこで本研究ではワイヤ速度$\bm{\dot{l}}$(モータに取り付けたエンコーダから取得)に対するフィードバックを行っている. 
速度フィードバック項を含む最終的な目標ワイヤ張力$\bm{f}_{\mathrm{ref}}^{\mathrm{final}}$は以下の式から求まる.
\begin{align}
  \label{eq:wire-tension}
  \bm{f}_{\mathrm{ref}}^{\mathrm{final}} = \bm{K}_{\mathrm{v}} (\bm{\dot{l}}_{\mathrm{ref}} - \bm{\dot{l}}) + \bm{f}_{\mathrm{ref}}
\end{align}
ただし$\bm{K}_{\mathrm{v}}$は速度のフィードバックゲイン行列, $\bm{\dot{l}}_{\mathrm{ref}}$は目標ワイヤ速度である.





}%

\begin{figure}[t]
  \centering
  \includegraphics[width=1.0\columnwidth]{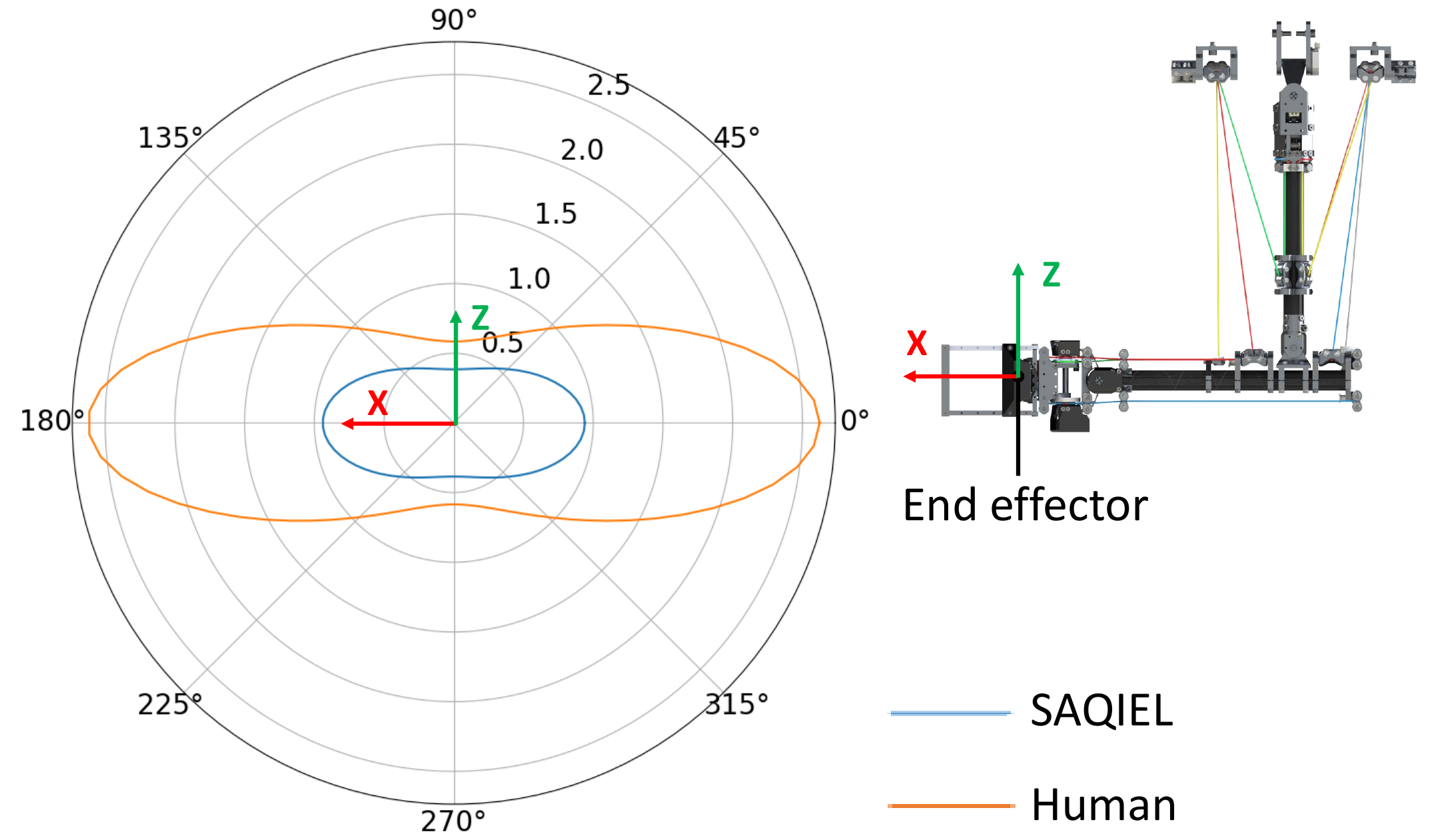}
  \vspace{-2ex}
  \caption{Effective mass of SAQIEL.}
  \label{figure:effectivemass}
  \vspace{-2ex}
\end{figure}

\section{Experiments} \label{sec:experiments}
\subsection{Analysis of Effective Mass} \label{subsec:mass-analysys}
\switchlanguage%
{%

The effective mass of a manipulator at a certain contact point refers to the perceived mass experienced when the manipulator makes contact \cite{khatib1995inertial}. 
As shown in Eq. \eqref{eq:fext_max}, a smaller effective mass leads to lower contact forces during collisions, enabling high-speed operation while maintaining safety \cite{Haddadin2015}. 
In this section, we numerically calculate the effective mass of the end effector of SAQIEL to verify its safety.

In \figref{figure:effectivemass}, the computed results of the effective mass for SAQIEL and the human arm in the $xz$ plane are depicted. 
The maximum effective mass of SAQIEL (excluding the motor rotor inertia) is \SI{0.94}{\kilogram}. 
This is approximately one-third of the maximum effective mass of the human arm (\SI{2.6}{\kilogram}), highlighting its lightweight nature. 
It should be noted that the calculation of the human effective mass employs a URDF model \cite{latella2019human-gazebo} of a human with the same upper limb length as SAQIEL.

Prior studies aiming to reduce effective mass, similar to our research, include the 7-DoF manipulators lims1 \cite{kim2017anthropomorphic} and lims2 \cite{song2018lims2}, which consolidated motors for elbow and wrist actuation in the upper arm. 
Lims1 had a maximum effective mass of \SI{1.5}{\kilogram}, while lims2 had a maximum effective mass of \SI{2.1}{\kilogram}.
In comparison, SAQIEL's effective mass is approximately 2/3 to 1/2 that of these prior studies. 
This demonstrates the significant contribution of the passive 3D wire aligners to the lightweight design of the manipulator.
}%
{%
  ある接触点におけるマニュピレータの有効質量とは, そのマニュピレータと接触したときに感じる賞味の質量を指す\cite{khatib1995inertial}.
  式\eqref{eq:fext_max}で示したとおり有効質量が小さいほど衝突時の接触力が小さくなり, 安全性を確保しつつ高速に動作することが可能となる\cite{Haddadin2015}.
  本節では, SAQIELの安全性を検証するためにエンドエフェクタの有効質量を数値計算で求める.
  
  \figref{figure:effectivemass}にxz平面におけるSAQIELおよび人間の腕部の有効質量の計算結果を示す.
  SAQIELの最大有効質量(モータのロータ慣性を除く)は\SI{0.94}{\kilogram}である.
  これは人間の腕部の最大有効質量\SI{2.6}{\kilogram}の約1/3であり非常に軽量であることが分かる.
  なお人間の有効質量の計算には, SAQIELと等しい上肢長の人間のURDFモデル\cite{latella2019human-gazebo}を用いた.
  
  本研究と同様に有効質量の低減を目指した先行研究として, 上腕部に肘・手首駆動用のモータを集約した7自由度マニュピレータlims1\cite{kim2017anthropomorphic}およびlims2\cite{song2018lims2}が挙げられる.
  Lims1の最大有効質量は\SI{1.5}{\kilogram}であり, lims2の最大有効質量は\SI{2.1}{\kilogram}である.
  SAQIELの有効質量はこれらの先行研究の2/3, 1/2程度となっており, 受動3次元ワイヤ整列装置がマニュピレータの軽量化に大きく貢献することが確認できる.
}

\begin{figure}[t]
  \centering
  \includegraphics[width=1.0\columnwidth]{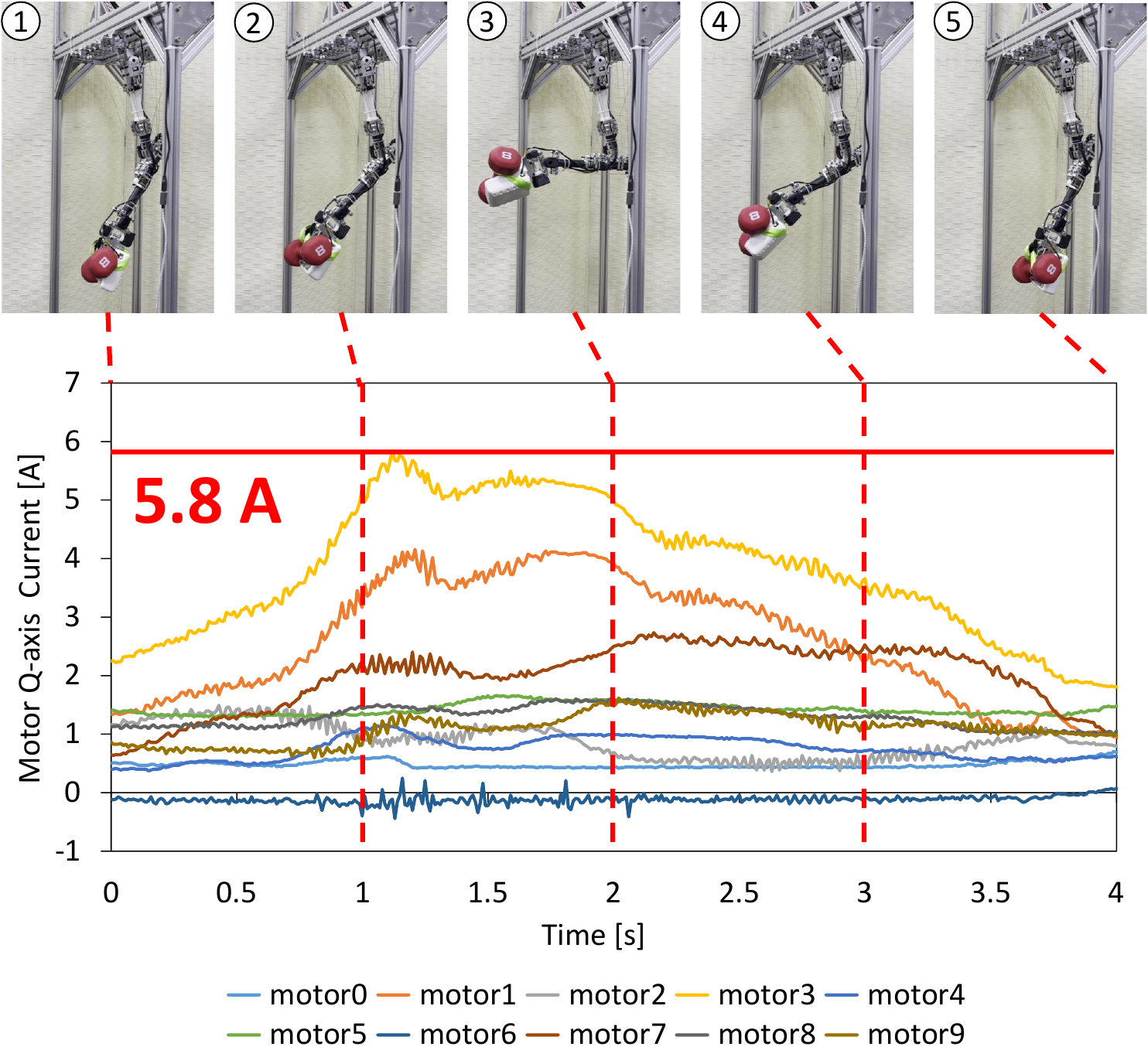}
  \vspace{-2ex}
  \caption{Palyload test with \SI{3.74}{\kilogram} weight. Maximum motor q-axis current is \SI{5.8}{\ampere} (power invariant transformation).}
  \label{figure:payloadexperiment}
  \vspace{-1ex}
\end{figure}

    

\subsection{Payload Test} \label{subsec:payload-exp}
\switchlanguage%
{%
To demonstrate SAQIEL's capability to exert sufficient force, a lifting motion test with a heavy payload is conducted. 
The sequential images of the test and the q-axis current values (power invariant transformation) of each motor are presented in \figref{figure:payloadexperiment}.

In this test, a weight of \SI{3.7}{\kilogram} is fixed to the end effector, and the elbow pitch joint is moved from \SI{-44}{\degree} to \SI{20}{\degree}. 
During this movement, the motors are subjected to q-axis currents of up to approximately \SI{5.8}{\ampere}, which is about half of the allowable peak current of \SI{10}{\ampere}. 
This test confirms that SAQIEL is both lightweight and flexible while providing sufficient actuating force.
}%
{%
  SAQIELが十分な力を発揮できることを示すため, 重量物の持ち上げ動作試験を行う.
  試験の連続写真と各モータのq軸電流値(絶対変換)を\figref{figure:payloadexperiment}に示す.

  本試験では\SI{3.7}{\kilogram}のおもりをエンドエフェクタに固定した状態で, elbow pitch関節を\SI{-44}{\degree}$\sim$\SI{20}{\degree}まで動かした.
  このときモータには最大で\SI{5.8}{\ampere}程度の電流が印加されており, これは許容ピーク電流\SI{10}{\ampere}の半分程度である.
  この試験からSAQIELが軽量・柔軟でありながら十分な発揮力を有していることが確認できる.
}%

\subsection{High Speed Motion Test} \label{subsec:speed-exp}
\begin{figure}[t]
  \centering
  \includegraphics[width=0.8\columnwidth]{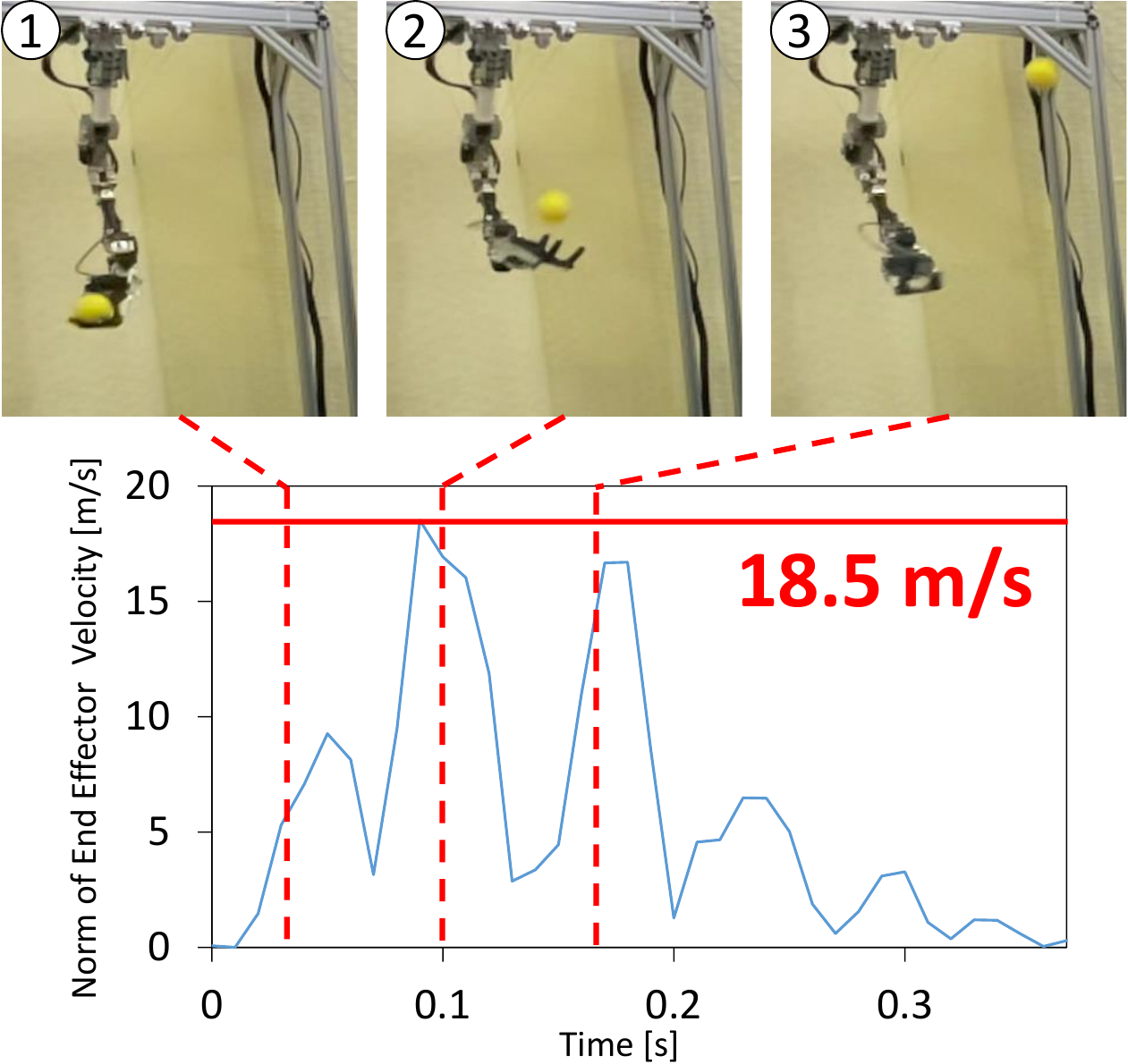}
  \vspace{-1ex}
  \caption{High speed motion test. SAQIEL throw a ball with the end effector speed of \SI{18.5}{\metre/\second}}
  \label{figure:highspeedexperiment}
  \vspace{-2ex}
\end{figure}

\switchlanguage%
{%
To demonstrate SAQIEL's capability for high-speed operation, a ball throwing experiment is conducted. 
Sequential images of the experiment along with the end effector velocity are depicted in \figref{figure:highspeedexperiment}.

In this experiment, a \SI{31}{\gram} ball is thrown from the end effector. 
Initially, the ball is gripped by three fingers (rigid rods without moving parts) on the end effector. 
By providing a desired trajectory to SAQIEL, both the end effector and the ball are accelerated. 
As the trajectory's end point is approached, the end effector decelerates, causing the ball to be propelled by its inertia.
 As explained in \secref{sec:controller}, trajectory tracking is achieved through position feedback from joint encoders and wire velocity feedback from motor encoders.

The maximum distance the ball travels is approximately \SI{4.8}{\metre}. 
Furthermore, the maximum end-effector velocity was approximately \SI{18.5}{\metre/\second}. 
This value significantly surpasses the operational speed of \SI{1}{\metre/\second} for the wave gear-driven collaborative robot \cite{ur3e2023} 
and exceeds the \SI{5.35}{\metre/\second} speed of other lightweight wire-driven manipulator \cite{kim2017anthropomorphic}. 
This extreme end-effector velocity is achieved by utilizing the lightweight and flexible characteristics inherent in the coupled tendon-driven system and vigorous swinging of the wrist.
This experiment confirms that due to its low effective mass, SAQIEL is capable of generating substantial end effector accelerations, making it well-suited for high-speed operations.
}%
{%
  SAQIELが高速に動作可能であることを示すため, ボールを投擲する実験を行う.
  実験の連続写真と手先速度を\figref{figure:highspeedexperiment}に示す.

  本試験では\SI{31}{\gram}のボールをエンドエフェクタから投擲する.
  最初ボールはエンドエフェクタ上の3本の指(可動部のない剛体の棒)に挟まっている.
  SAQIELに目標軌道を与えることでエンドエフェクタとボールを加速させる.
  そして軌道の終端でエンドエフェクタが減速し, ボールが慣性で飛び出す.
  \secref{sec:controller}で述べたとおり, 目標軌道への追従は関節エンコーダを用いた位置フィードバックとモータのエンコーダを用いたワイヤ速度フィードバックによって行っている.

  ボールの最大飛距離は約\SI{4.8}{\metre}程度である.
  また, 最大手先速度は約\SI{18.5}{\metre/\second}であった.
  これはwave gear駆動の協働ロボット\cite{ur3e2023}の動作速度\SI{1}{\metre/\second}や, 他の軽量ワイヤ駆動マニュピレータ\cite{kim2017anthropomorphic}の動作速度\SI{5.35}{\metre}{/\second}と比較して優位に大きい値である.
  拮抗ワイヤ駆動特有のリンクの軽量性と柔軟性を活用し手首を大きく振動させることで, この手先速度を達成している.
  この実験から, SAQIELは有効質量が小さいため大きな手先加速度を発揮可能であり, 高速動作に適していることが確認できる.
}%

\subsection{Accuracy Test} \label{subsec:accuracy-exp}
\begin{figure}[t]
  \centering
  \includegraphics[width=1.0\columnwidth]{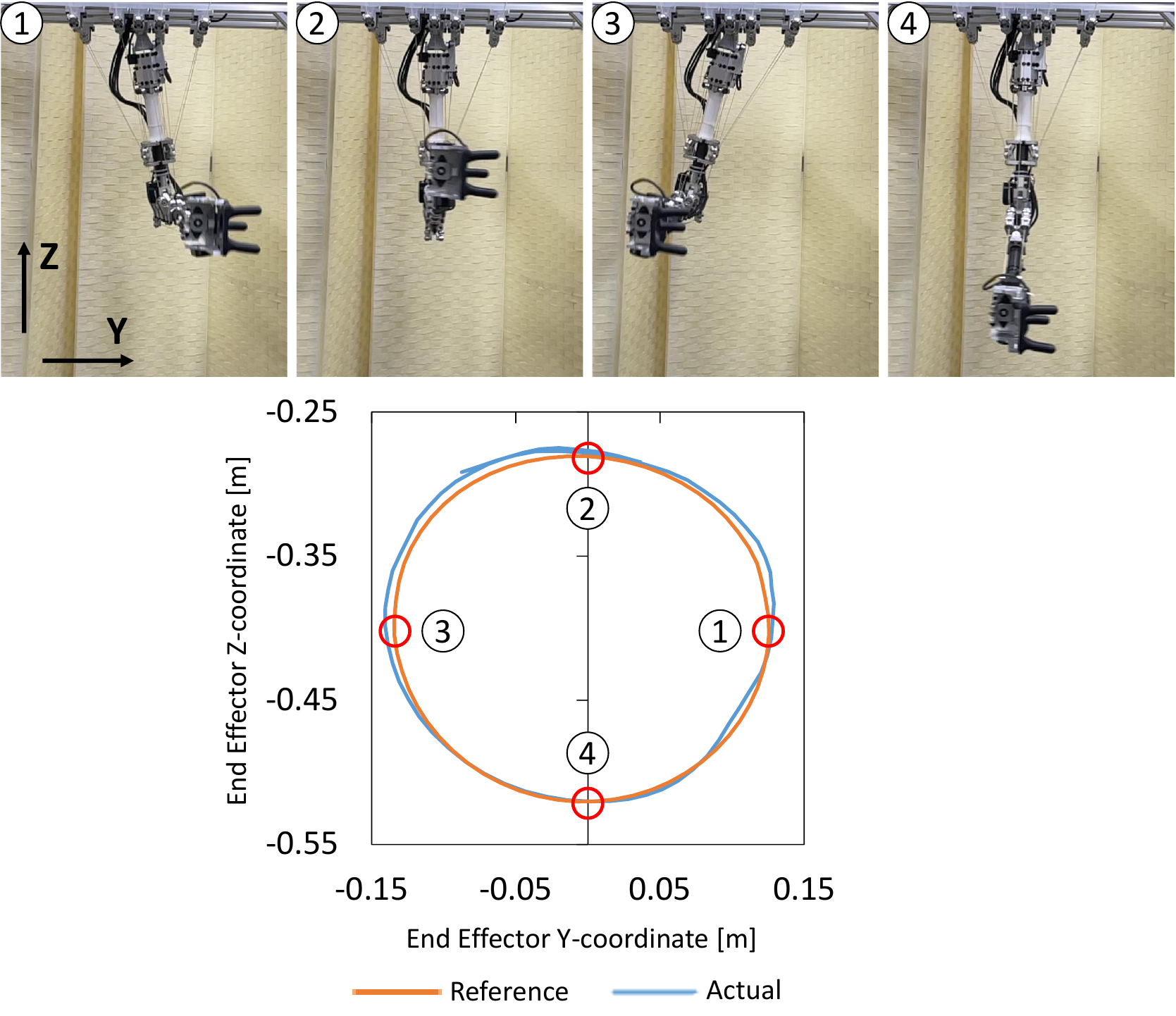}
  \vspace{-2ex}
  \caption{Cicular trajectory (diameter:\SI{0.25}{\metre}, period:\SI{0.6}{\second}) following experiment.}
  \label{figure:accuracyepperiment}
  \vspace{-0.5ex}
\end{figure}

\switchlanguage%
{%
To demonstrate SAQIEL's position control capability, a high-speed trajectory tracking experiment is conducted. 
Sequential images of the experiment and the trajectory of the end effector position are illustrated in \figref{figure:accuracyepperiment}.

In this experiment, the end effector is made to track a circular trajectory (diameter of \SI{0.25}{\metre}, period of \SI{0.6}{\second}). 
The maximum position error of the end effector during this trajectory tracking is \SI{11}{\milli\metre}. 

The primary causes of end-effector position errors include cogging torque in the motor, friction between the wire and pulley, and the elasticity of the wire, which result in discrepancies between the commanded tension and the actual tension. 
Incorporating these factors into the control model can be expected to reduce the error between commanded tension and actual tension, thus improving end-effector positioning accuracy.
}%
{%
  SAQIELの位置制御能力を示すため, 高速軌道追従実験を行う.
  試験の連続写真と手先位置の軌跡を\figref{figure:accuracyepperiment}に示す.

  本試験ではエンドエフェクタを円軌道(直径\SI{0.25}{\metre}, 周期\SI{0.6}{\second})に追従させた.
  このときの最大手先位置誤差は\SI{11}{\milli\metre}である.
  
  手先位置誤差の主な発生原因としては, モータのコギングトルク, ワイヤ-プーリ間の摩擦, ワイヤの弾性などによって指令張力と実際の張力に誤差が生じていることが挙げられる.
  これらを制御モデルに組み込み指令張力と実張力の誤差を低減することで手先位置決め精度の向上が期待できる.
}%

\begin{figure}[t]
  \centering
  \includegraphics[width=1.0\columnwidth]{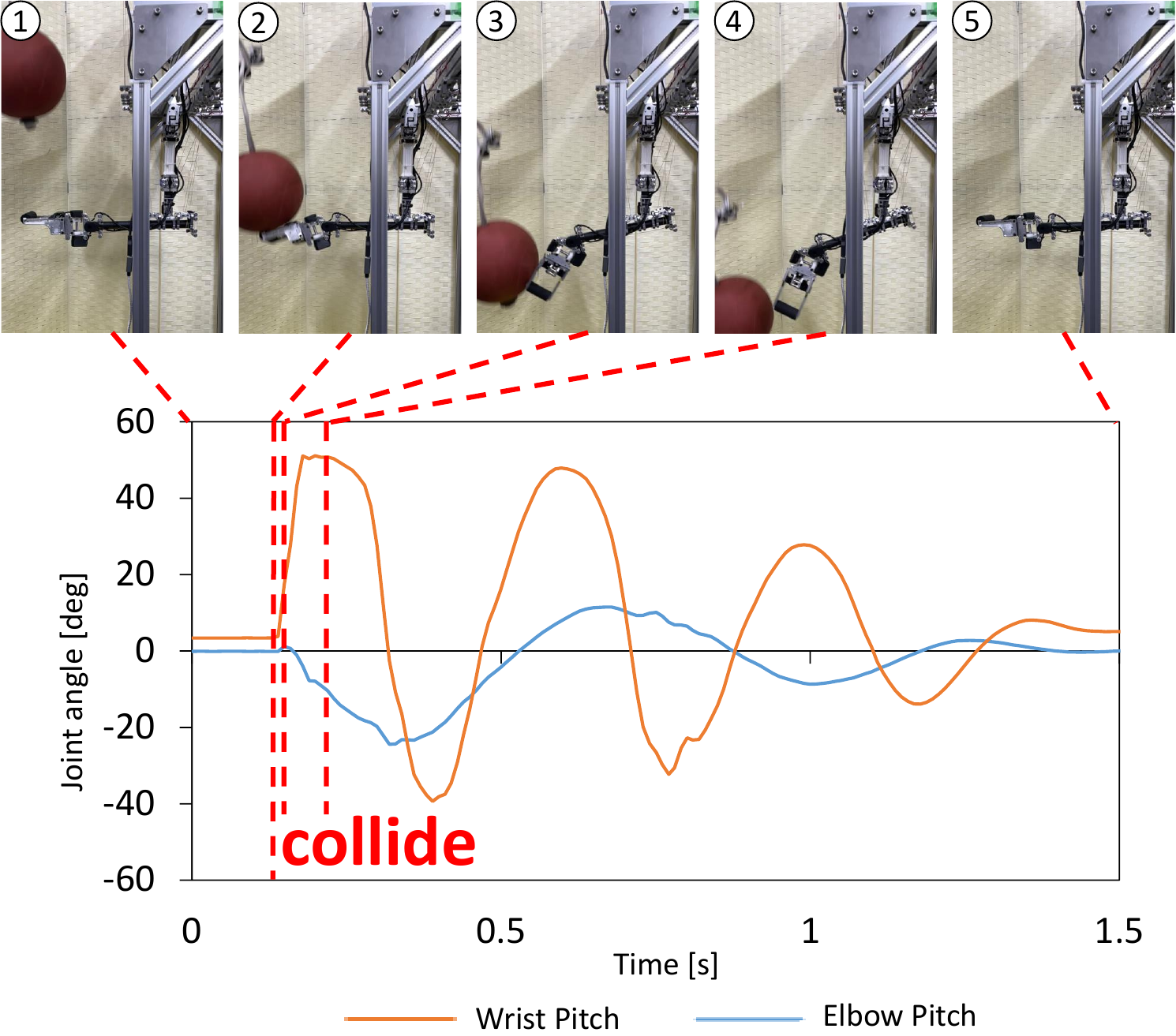}
  \vspace{-2ex}
  \caption{Passive collision test. Dropping a \SI{5}{\kilogram} weight from a height of 1m and inducing collision with SAQIEL.}
  \label{figure:ballcollisionexperiment}
  \vspace{-1ex}
\end{figure}

\subsection{Passive Collision Test} \label{subsec:collision-exp}
\switchlanguage%
{%
To assess the response of SAQIEL when subjected to external impact, a collision experiment is conducted by striking a weight against SAQIEL. 
Sequential images of the experiment and the changes in wrist pitch and elbow pitch joint angles are presented in \figref{figure:ballcollisionexperiment}.

In this experiment, a \SI{5}{\kilogram} ball is dropped from a height of \SI{1}{\metre} onto the SAQIEL, which is under position control.
The velocity of the weight at the moment of impact is estimated to be approximately \SI{4.4}{\metre/\second} based on the fall distance. 
To enhance flexibility, the gains for joint angle feedback and wire velocity feedback are set to about one-tenth of those in \secref{subsec:accuracy-exp}.

From \figref{figure:ballcollisionexperiment}, it can be observed that after the collision, the wrist pitch joint angle shifts by \SI{50}{\degree} within \SI{0.05}{\second}, and the elbow pitch joint angle shifts by \SI{-24}{\degree} within \SI{0.17}{\second}. 
Responding quickly and flexibly to external impacts in this manner is challenging for robots with high friction in their drive systems or substantial moving part masses. 
This experiment confirms that SAQIEL can respond softly and safely to external impacts.
}%
{%
  SAQIELに外部から撃力を加えたときの応答を確認するため, SAQIELにおもりを衝突させる.
  実験の連続写真とwrist pitch及びElbow pitchの関節角度の変化を\figref{figure:ballcollisionexperiment}に示す.

  本実験では位置制御を行っているSAQIELに\SI{5}{\kilogram}のボールを\SI{1}{\metre}上空から落とし衝突させる.
  衝突時のおもりの速度は約\SI{4.4}{\metre/\second} (落下距離から推定)である.
  柔軟性を上げるために関節角度フィードバックおよびワイヤ速度フィードバックのゲインを\secref{subsec:accuracy-exp}の1/10程度にしている.

  \figref{figure:ballcollisionexperiment}から, 衝突後にWrist pitch関節が\SI{0.05}{\second}で\SI{50}{\degree}, Elbow pitch関節が\SI{0.17}{\second}で\SI{-24}{\degree}変位していることが分かる.
  撃力に対してこのように素早く柔軟に応答することは, 駆動系の摩擦が大きいロボットや動作部重量が大きいロボットには困難である.
  この実験からSAQIELが外部からの撃力に対して柔らかく安全に対応可能であることが確認できた.
}%

\begin{figure}[t]
  \centering
  \includegraphics[width=1.0\columnwidth]{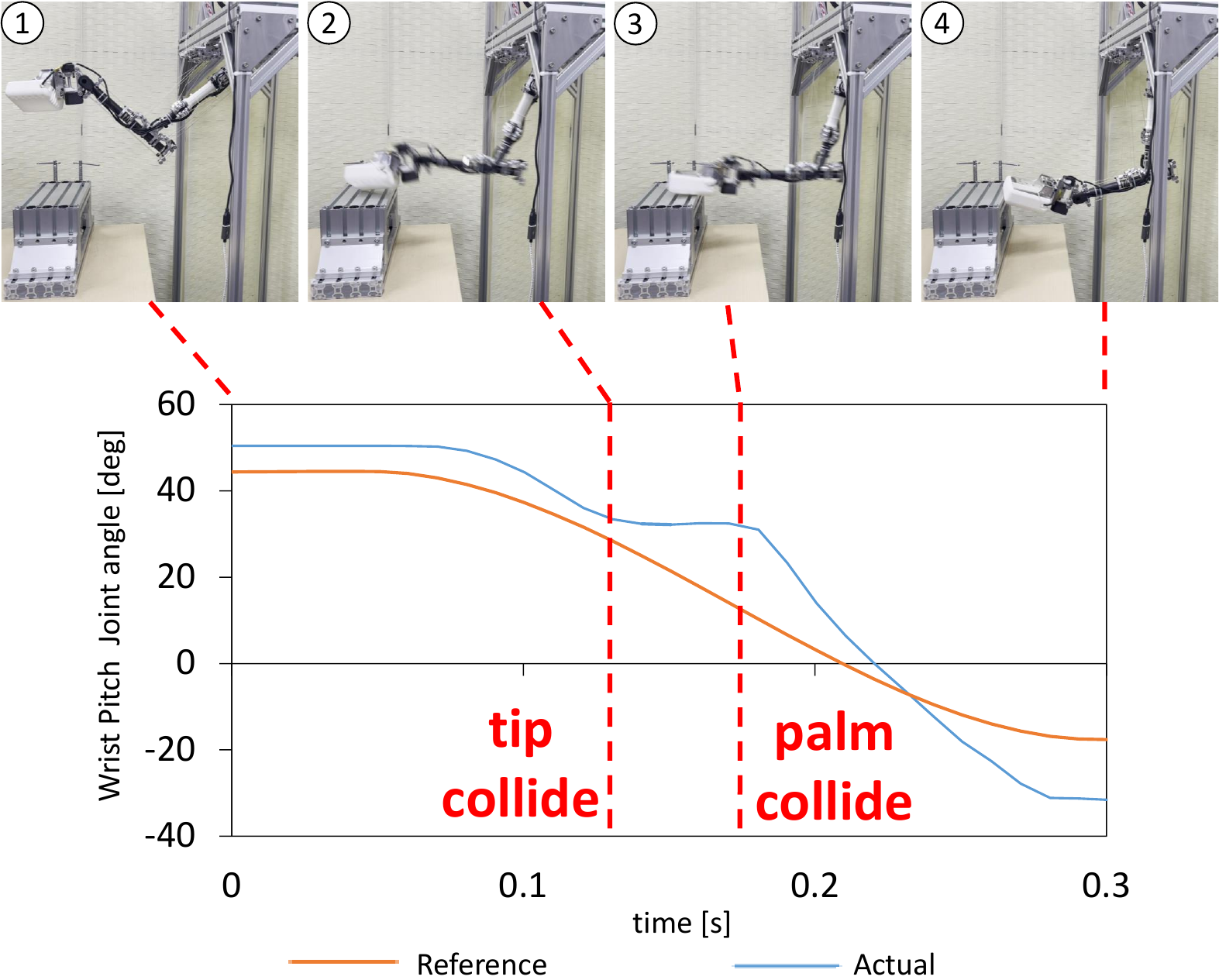}
  \vspace{-2ex}
  \caption{Active collision test. SAQIEL impacts an aluminium frame.}
  \label{figure:tableimpactexperiment}
  \vspace{-1ex}
\end{figure}

\subsection{Active Collision Test} \label{subsec:active-collision-exp}
\switchlanguage%
{%
To assess the response of SAQIEL when colliding with the environment during operation, an experiment is conducted where SAQIEL strikes an aluminum frame. 
Sequential images of the experiment and changes in wrist pitch joint angle are presented in \figref{figure:tableimpactexperiment}.

In this experiment, similar to \secref{subsec:accuracy-exp}, SAQIEL follows a target trajectory. 
The collision between SAQIEL and the aluminum frame is set to occur at the lower end of the target trajectory. 
The collision portion of SAQIEL is made of rigid components created using a 3D printer, without the use of elastic materials such as rubber or springs for impact absorption.

From the sequential images in \figref{figure:tableimpactexperiment}, it can be observed that SAQIEL's tip and palm sequentially come into contact with the aluminum frame. 
Furthermore, from the plot of the wrist pitch joint angle, it can be confirmed that within \SI{0.1}{\second} after the collision, the end effector moves along the surface of the aluminum frame.
These results indicate that SAQIEL is capable of flexible environmental contact even during operation.

Performing such intense contact with the environment on a typical industrial manipulator can pose a risk of damaging both the environment and the robot. 
However, in this experiment, neither SAQIEL nor the aluminum frame exhibited any damage. 
This suggests that lightweight and flexible structures like SAQIEL's have the potential to enhance the safety of robots by mitigating the risk of damage in such environmental interactions.
}%
{
  SAQIELが動作中に環境に衝突した際の応答を確認するため, SAQIELでアルミフレームを打撃する.
  実験の連続写真とwrist pitchの関節角度の変化を\figref{figure:tableimpactexperiment}に示す.

  本実験では\secref{subsec:accuracy-exp}と同様に, SAQIELに目標軌道を追従させている.
  目標軌道の下端でSAQIELとアルミフレームが衝突するよう調節した.
  SAQIELの衝突部は3Dプリンタ製の剛な部品であり, ゴムやバネなどの弾性体による衝撃吸収は行っていない.

  \figref{figure:tableimpactexperiment}の連続写真から, SAQIELの手先と手の平が順にアルミフレームと接触していることが分かる.
  さらにwrist pitch関節角度のプロットから, 手先が衝突してから\SI{0.1}{\second}後には手先がアルミフレーム表面に沿って動いていることも確認できる.
  これらの結果から, SAQIELは動作中においても環境との柔軟な接触が可能であることが確認できる.

  通常の産業用マニュピレータでこのような環境へ激しく接触する動作を行うと, 環境及びロボットを破損する危険性がある.
  一方で本実験ではSAQIELおよびアルミフレームの損傷は見られなかった.
  このことからSAQIELのような軽量・柔軟な構造がロボットの安全性を向上させることが期待できる.
}%

\section{Conclusion} \label{sec:conclusion}
\switchlanguage%
{%
In this study, we propose a lightweight and low-friction power transmission mechanism called the passive 3D wire aligner to achieve a lightweight, flexible, and safe manipulator. 
By using this approach to transmit actuator power from the root link, a simple and lightweight moving part can be created. 
Furthermore, as this method employs a wire and needle bearing-based power transmission, it ensures low friction due to the absence of sliding components.

To validate the performance of this approach, we developed a 7-DoF wire-driven manipulator named SAQIEL. 
Through numerical analysis, we confirmed that SAQIEL's effective mass is approximately two-thirds of that of conventional lightweight manipulators\cite{kim2017anthropomorphic}. 
Furthermore, we verified its capability to handle sufficient payloads, achieve maximum velocities, maintain position control, and exhibit flexibility for various tasks.
Additionally, tracking experiments of end-effector trajectories have been conducted, indicating that friction between the wire and pulley, as well as the elasticity of the wire, are the causes of end-effector position errors.

Future challenges include measuring and enhancing the transmission efficiency of the passive 3D wire aligner.
By evaluating the transmission efficiency in different wire materials and pulley shapes, further performance improvements are anticipated. 
Moreover, advanced tasks such as introducing a parallel link structure for wire interference avoidance and exploring applications in leg mechanisms can be considered.
Furthermore, by optimizing wire arrangements and increasing the implementation density of motor modules and 3D passive wire aligners, it becomes possible to realize smaller and lighter root links. 
This opens up the potential for applications in humanoid robots and wheeled robots.
}%
{%


本研究では軽量・柔軟で安全なマニュピレータの実現のために, 受動3次元ワイヤ整列装置という軽量・低摩擦な動力伝達機構を提案した.
本手法を用いてルートリンクからアクチュエータ動力を伝達することで, 簡素で軽量な動作部を作成可能である.
さらに本手法はワイヤとニードルベアリングによる動力伝達方式であるため, 摺動部が無く低摩擦である.

本手法の性能を検証するため, 7自由度ワイヤ干渉駆動マニュピレータSAQIELを作成した.
数値解析を通して, SAQIELの有効質量が従来の軽量マニュピレータ\cite{kim2017anthropomorphic}の2/3程度になっていることを確認した.
さらにタスクに十分なペイロード, 最大速度, 柔軟性を有していることも確認した.
また手先軌道への追従実験を行い, ワイヤ-プーリ間の摩擦やワイヤの弾性が手先位置誤差の原因となっている可能性を指摘した.

今後の課題としては, 受動3次元ワイヤ整列装置の伝達効率の測定および向上が挙げられる.
様々なワイヤ素材やプーリ形状における伝達効率を検証することでさらなる性能の向上が見込まれる.
また発展的な課題としてワイヤ干渉回避のためのパラレルリンク構造の導入, 脚部への応用などが考えられる.
さらにワイヤ配置を工夫しモータモジュールや3次元受動ワイヤ整列装置の実装密度を高めることで, より小型軽量なルートリンクを実現しヒューマノイドや台車型ロボットへの応用も可能である.

}%

{
  \bibliographystyle{IEEEtran}
  \bibliography{main}
}

\end{document}